\documentclass[5p, times]{elsarticle}

\usepackage{lineno}
\usepackage{booktabs}
\usepackage{multirow}
\usepackage{algorithmic,algorithm}
\usepackage{graphicx}
\usepackage{array}
\usepackage{xcolor}
\usepackage{soul}
\usepackage{todonotes}
\usepackage[caption=false,font=footnotesize]{subfig}
\usepackage[colorlinks,hidelinks]{hyperref}
\usepackage[capitalize,noabbrev,nameinlink]{cleveref}

\modulolinenumbers[1]

\journal{Solar Energy}
\renewcommand*{\today}{May 15, 2021}

\biboptions{authoryear}

\begin{document}

\begin{frontmatter}

\title{\LARGE Segmentation of Cell-Level Anomalies in Electroluminescence Images of Photovoltaic Modules}

\author[VICOMTECH]{Urtzi Otamendi }\corref{correspondingauthor}
\cortext[correspondingauthor]{Corresponding author}
\ead{uotamendi@vicomtech.org}

\author[VICOMTECH,TECNUN]{Iñigo Martinez}

\author[VICOMTECH]{Marco Quartulli}

\author[VICOMTECH]{Igor G. Olaizola}

\author[TECNUN,ICDIA]{Elisabeth Viles}

\author[TECNALIA]{Werther Cambarau}



\address[VICOMTECH]{Vicomtech Foundation, Basque Research and Technology Alliance (BRTA), Donostia-San Sebastián 20009, Spain}

\address[TECNUN]{TECNUN School of Engineering, University of Navarra, Donostia-San Sebastián 20018, Spain}

\address[ICDIA]{Institute of Data Science and Artificial Intelligence, University of Navarra, Pamplona 31009, Spain}

\address[TECNALIA]{Tecnalia Research \& Innovation, Basque Research and Technology Alliance (BRTA), Donostia-San Sebastián 20009, Spain}

\begin{abstract}
In the operation \& maintenance (O\&M) of  photovoltaic (PV) plants, the early identification of failures has become crucial to maintain productivity and prolong components' life. 
Of all defects, cell-level anomalies can lead to serious failures and may affect surrounding PV modules in the long run.
These fine defects are usually captured with high spatial resolution electroluminescence (EL) imaging. The difficulty of acquiring such images has limited the availability of data. For this work, multiple data resources and augmentation techniques have been used to surpass this limitation. 
Current state-of-the-art detection methods extract barely low-level information from individual PV cell images, and their performance is conditioned by the available training data.
In this article, we propose an end-to-end deep learning pipeline that detects, locates and segments cell-level anomalies from entire photovoltaic modules via EL images. The proposed modular pipeline combines three deep learning techniques: 1. object detection (modified \textit{Faster-RNN}), 2. image classification (\textit{EfficientNet}) and 3. weakly supervised segmentation (\textit{autoencoder}). 
The modular nature of the pipeline allows to upgrade the deep learning models to the further improvements in the state-of-the-art and also extend the pipeline towards new functionalities.
\end{abstract}


\end{frontmatter}

\begin{sloppypar}
\section{Introduction}
\label{section:introduction}

Solar energy has dominated the expansion of renewable energy capacity in recent years. The installation of photovoltaic energy has increased since 2010, when manufacturing prices started to decrease, driving more than 110 countries to invest in solar energy \citep{IEA_road}. As a result, record-level PV capacity growth has been headlining renewable energy news over the last years. Recent studies assert that the capacity of photovoltaic solar energy surpassed 627 GW in 2019 \citep{IEA_2019}, and the \textit{IEA}'s latest 5-year forecast shows that the total capacity will reach 1209 GW in 2024 \citep{IEA_forecast}. 

Besides, the photovoltaic services market of O\&M and installations is estimated to grow by 10.81 billion dollars during the period 2019-2023, which represents a growth of 16\% per year \citep{BusinessWire}. In this regard, the digitization of O\&M is presumed to reduce the costs while boosting the performance of PV installations. The early identification of panel failures and deterioration is crucial to avoid a reduction in production efficiency (productivity) and prolong the life of the components \citep{powerloss3, powerloss4}.

A photovoltaic (PV) panel can have different types of anomalies depending on the element it affects and the loss of productivity it causes. Major anomalies such as \textit{panel degradation}, \textit{electrical disconnection} or \textit{hot spots}, cause heat emission under abnormal functioning, and thus the damaged areas can be easily revealed using infrared imagery (IR). 
On the other hand, cell level anomalies -- such as \textit{material defects, fingerprint marks, micro-cracks or electrically isolated parts} \citep{el_extr_intr, norma} -- may not directly affect the functioning of the panel but in the long run may lead to the appearance of serious failures \citep{powerloss, powerloss2}, also affecting surrounding PV modules \citep{solarefficiency}.

Cell level anomalies cause slight temperature differences between non-defective and defective areas. Various studies \citep{el, norma, deitsch_2019} have shown that these types of anomalies are harder to detect using optical or infrared imagery, so electroluminescence (EL) imaging is used instead. The operation of a PV cell is to absorb light and convert it into electricity. The reciprocity principle allows the opposite, that is, applying a direct current into the PV module and measuring the infrared photoemission with a special camera configuration \citep{elpv1}. EL imaging provides insight into micro-cracks and other defects within the cell material, a pivotal information for O\&M. The emitted light has a peak wavelength of $1150 nm$ \citep{electroluminescence_peak}, which is very appropriate to reveal the most slight and subtle anomalies.


\begin{figure*}[!hbt]
\centering
 \includegraphics[width=0.8\linewidth]{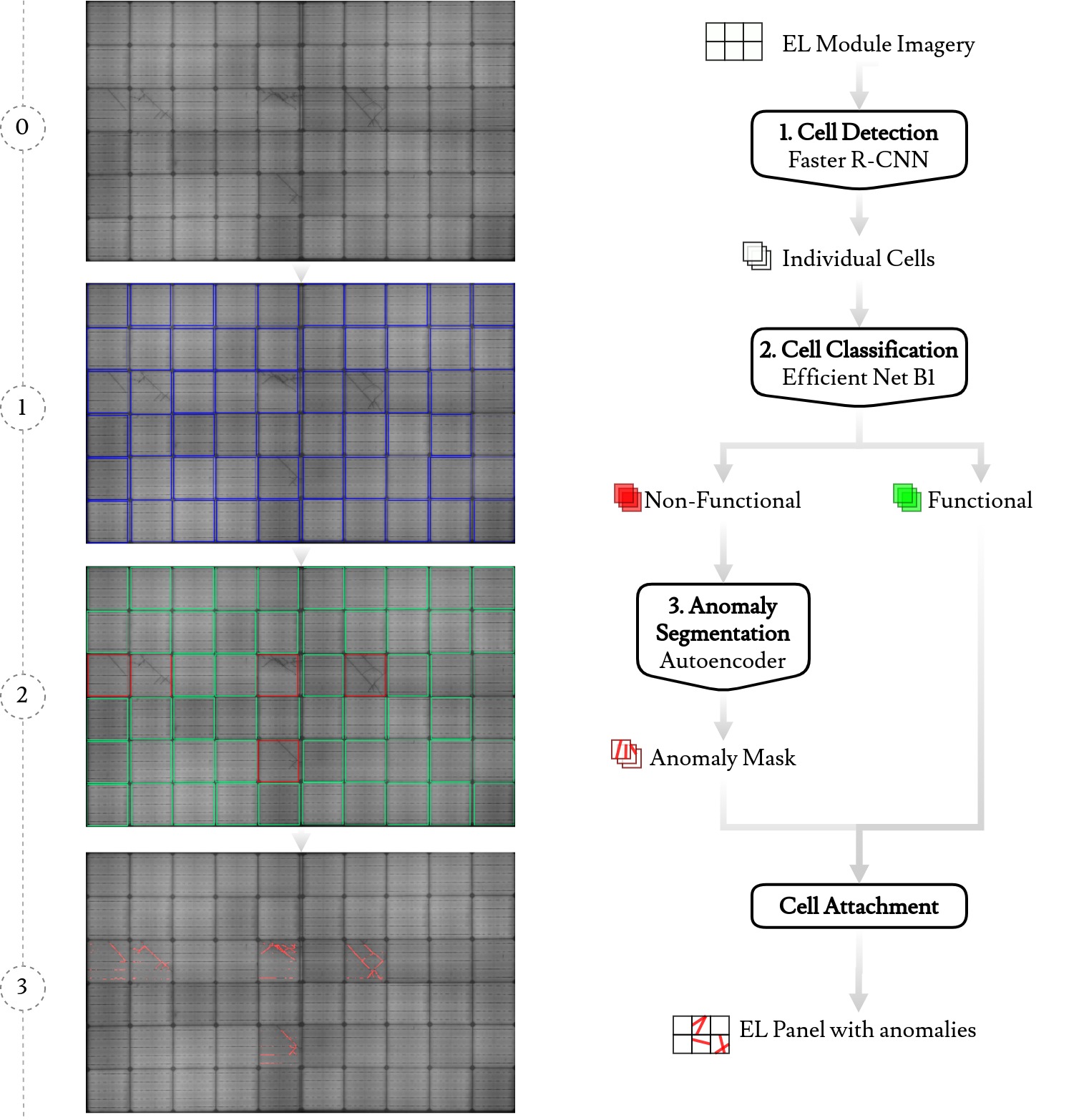}
\caption{Workflow of the proposed segmentation architecture, from the input EL image to the output segmented image, through the four processing modules.
The images on the left represent (from top to bottom): 0. Original EL module; 1. Bounding box of each cell (blue); 2. Classification of each cell as non-defective (green) or defective (red); 3. Segmentation of the defective zones (red)}
\label{fig:architecture_el}
\end{figure*}
\vspace{-3pt}

In this sense, anomaly segmentation techniques are able to detect and locate with precision the conflicting elements in an image. The process of partitioning an image into multiple segments (sets of pixels) requires a high-level of comprehension of the domain and image itself. In this regard, a number of recent contributions in the field of anomaly segmentation in images have used deep learning techniques. Deep learning models are able to learn and recognize complex and abstract patterns in images like no other machine learning model \citep{Lecun2015436, Litjens201760}. 
Due to the capacity of these models to exploit such abstract patterns, their performance on the task of anomaly segmentation is currently state-of-the-art \citep{He2016770, Long20153431}. 

Even though research is being conducted to detect irregularities in PV panels using conventional machine learning techniques, there is a lack of methods that extract high-level features such as anomaly shape, contour, or semantically meaningful information. Hence, in this article a deep-learning pipeline is proposed to detect, locate and segment cell level anomalies from images of photovoltaic modules (\textit{see \cref{fig:architecture_el}}). To the best of our knowledge, this is the first work proposing a complete end-to-end solution for the detection and segmentation of cell-level anomalies in PV modules.

The rest of this paper is organized as follows.
Section 2 surveys the related work of anomaly detection in photovoltaic panels using EL imagery.
Section 3 introduces the main contributions of the article. 
Section 4 presents a detailed explanation of the proposed pipeline. 
Section 5 describes the EL dataset used for the training and validation tasks. 
Section 6 presents the results obtained from the application of the proposed pipeline to the datasets introduced in Section 5.
Finally, Section 7 contains the concluding remarks and outlines some areas for further research.

\section{Related work}
\label{section:related_work}

The detection of anomalies in photovoltaic panels has evolved from the early use of optical images to the recent adoption of more specific images such as multi-spectral, thermal, optical, etc. \citep{infrared1, infrared2, imagery3, imagery2}. Recently, new detection approaches have emerged that handle electroluminescence (EL) imagery, which allow the detection of slight anomalies. This section presents current approaches for the detection of anomalies using EL imagery.
 

One of the first attempts to use deep learning techniques with EL images on the domain of photovoltaic panels was introduced by \cite{deitsch_2018}. In this work, an automated segmentation method was proposed for the extraction of individual solar cells from EL images of PV modules. This method applies a bottom-up pipeline that exploits low-level edge features to progressively infer a high-level representation of the solar module and its cells. 

However, this approach is not robust to distinct variations of PV modules, such as the cell number, construction materials, type of cells, etc. It lacks the ability to capture abstract features among large and diverse data, which other methods, such as deep learning techniques, can provide.
It should also be noted that this kind of technique does not make any anomaly detection or location over the images. It exclusively focuses on the extraction of solar cells from PV panels, so it is rather a pre-processing technique.

In the interest of strengthening the state-of-the-art, \cite{deitsch_2019} continued making contributions to the field of anomaly detection in PV panels via EL imagery, publishing a method for automatic classification of defective photovoltaic module cells. This work introduces two novel approaches for the automatic detection of cell-level defects in a single image of a PV cell: a) a hardware-efficient approach that uses a Support Vector Machine (SVM) to classify hand-crafted features and b) a more hardware-demanding approach that uses an end-to-end deep Convolutional Neural Network (CNN) that runs on a Graphics Processing Unit (GPU). 
The application of an SVM implies the use of hand-crafted features, making the classifier less robust to the domain's intrinsic variations and therefore, less scalable. In turn, using a deep CNN implies a greater abstraction level and hence, more robustness and scalability. According to the tests carried out by the researchers, the approach of deep CNNs is more accurate than the SVM. 

The work of \cite{deitsch_2019} reveals with strong evidence the benefits of using Deep Learning techniques to classify anomalies. \cite{tang2020deep} contributed to the state-of-the-art publishing a comparison of the performance of various models: VGG16, ResNet50, Inception V3, and MobileNet. In this work, the authors generated the experimentation dataset using an approach combining traditional image processing technology and Generative Adversarial Network (GAN) characteristic. Although the classification is focused on the defect type instead of detecting non-defective and defective cells, the obtained average precision achieved state-of-the-art results.

In these latter approaches \citep{deitsch_2019, tang2020deep}, the deep learning architectures were not created specifically for anomaly detection in solar panels, but for classification challenges, where they achieved state-of-the-art performance. \cite{akram2019cnn} presented a novel approach using a light convolutional neural network architecture for recognizing defects that achieved the state-of-the-art. In this case, the model classifies a cell as defective or non-defective instead of providing more detailed information as in the previous case.

On the same innovative line, \cite{chen2019detection} proposed a novel model called SEF-CNN to tackle anomaly detection by exploring the function of traditional filters in deep learning approaches. This method filters the images using steerable evidence filters (SEF), making the features extracted by the convolutional neural network more discriminative and robust.

These approaches only provide the defective level of the solar cell but do not disclose details of these defects. \cite{mayr} tried to solve this problem by introducing a weakly supervised strategy for the segmentation of cracks on solar panels. The proposed strategy uses ResNet-50 \citep{resnet}, one of the most used CNN, with some modifications to derive a segmentation from the network's activation maps. The main contribution of the mentioned paper is the application of \textbf{\textit{$L_{p}$ }} normalization to aggregate the activation maps into single scores for classification.
However, this method obtains the segmentation of the anomaly from the activation map, making the mask less precise and concrete. \cite{rahman2020defects} proposed an algorithm that leverages the advantage of multi attention networks to efﬁciently extract the most important features and neglect the nonessential ones. They incorporate a modified U-net called multi attention U-net (MAU-Net) which requires well-annotated data to be trained but can segment and detect various complex defect masks correctly. The U-Net architecture was also exploited by \cite{balzategui2020defect}.

\section{Contributions overview}
\label{section:contributions_overview}

Based on the reviewed state-of-the-art, the number of publications related to the segmentation of anomalies in PV cell images is very limited. The mentioned approaches focus entirely on the anomaly detection of individual, independent PV cell images. As a matter of fact, data collection procedures generate images of entire PV panels, which are comprised of several cells. To apply these methods directly to PV panel images, a preprocessing step is necessary, such as the one proposed by \cite{deitsch_2018}.

In addition, the available datasets are limited to a specific type of cell, and there are few variations in terms of shape, material and structure. This lack of diversity and quantity of images has not prevented the application of deep learning techniques, but it has not allowed extracting the full potential from them, giving as a result less effective and less robust techniques.

In this article, the aforementioned gap in the literature is addressed, that is, the lack of methods that extract high-level anomaly information (localization) from PV panel imagery. Therefore, in this publication a novel pipeline is proposed to detect, locate and segment cell level anomalies from EL images of photovoltaic modules, extracting high-level information using various deep learning techniques.

\begin{figure}[t]
\centering
\includegraphics[width=\linewidth]{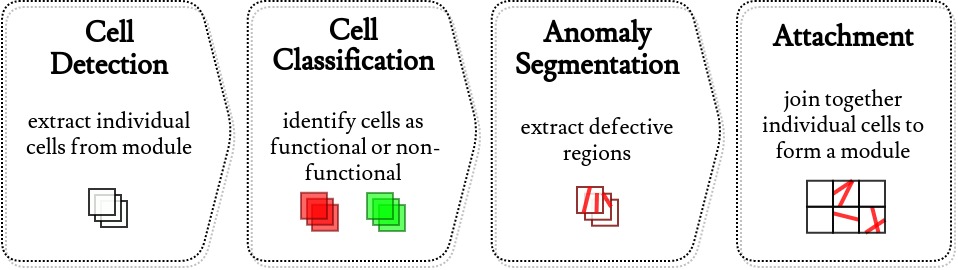}
\caption{Overview of the proposed pipeline divided on four processing modules: \textit{Detection}, \textit{Classification}, \textit{Segmentation} and  \textit{Attachment}.}
\label{fig:pipeline_el}
\end{figure}

The idea is to create a deep learning based end-to-end pipeline, divided into four adjacent processing units, creating a sequential workflow as illustrated in \cref{fig:pipeline_el}. The proposed end-to-end system is able to locate cell-level anomalies or defects of a solar module using an EL image.


The first module (\textit{cell detection}) takes an image of an entire PV module and extracts from it all the PV cells, detecting them one by one. Then, each cropped cell is processed on the second module (\textit{cell classification}) and is labeled as non-defective or defective. The cells that are classified as defective are further processed by the third module (\textit{failure segmentation}) and are segmented in order to locate the defective areas. Finally, the last module (\textit{attachment}) processes both the non-defective and defective segmented cells images, and attaches them to create a complete image of the PV panel, with all the defective zones detected. See \cref{fig:architecture_el} for an illustration of the proposed workflow. The first three units (\textit{detection, classification and segmentation}) make use of deep learning techniques to accomplish their task, and the attachment stage will be the only one to use a naive algorithm.

The development of the proposed pipeline can be a valuable contribution to the state-of-the-art of anomaly detection in photovoltaic modules, and to the sector working on early failure mechanisms and operation and maintenance of PV. 

In addition, the modularity of the pipeline allows the training of the three models simultaneously. The training will only require two datasets, one for cell detection and the other for cell classification and segmentation, as it is described in \cref{section:experimental_analysis_and_results}.

The proposed end-to-end system has been tested with polycrystalline-Si and monocrystalline-Si modules with full cell and half-cut cells of 3 and 5 busbars. Besides, data from other sources can be efficiently fed into the proposed deep learning models, thus improving their performance and allowing more types of anomalies to be located and segmented. Therefore, the pipeline can be adapted and extended to work with new PV cell technologies that may appear on the market, such as multi-busbar cells or bifacial cells.
Such advantage also applies to the segmentation model, which extracts the defective regions of a cell by encoding the distribution of non-defective cells in a compressed form. When the segmentation model is presented with a new cell anomaly that is absent in the training dataset, it will be still able to detect it, considering the model has retained a reference of what a non-defective cell is.

It should be noted that EL images usually require perspective distortion correction and lens distortion removal among other rectifications. These functions are not included in the proposed approach, so if required, it will be necessary to perform the rectification prior to the use of the pipeline.

In other respects, during the inference of the detection model, some outer areas of the cell may be omitted. To avoid losing the features of the outer part, an offset of 10\% of the cell size is applied to expand the inferred box. Even though the \cref{fig:architecture_el} shows the original boxes, the classification model receives the expanded boxes.
This post-processing technique allows minimizing the impact of the outer regions omitted by the detection model in subsequent stages of the pipeline.


\section{Methodology}
\label{section:methodology}

Having introduced the main architecture of the pipeline, in this section a thorough description of the selected methods for each module is presented. It is worth noting that the development of each module is independent of the rest.

\subsection{Detection module}
\label{section:detection_module}

The objective of the detection phase is to extract all PV cells from the PV panel image, without having to classify them among distinct classes. 
Object detection deep learning architectures are divided into two main classes. Two-stages detectors \citep{r-cnn,r-fcn} are based on region proposals, which means they generate regions of interest in the first stage and then process those regions for object classification and bounding-box regression. On the other hand, one-stage models \citep{yolo,ssd} target the task as a simple regression problem, learning class probabilities and bounding box coordinates.

Two-stage models have the highest accuracy rates in major object detection competitions, while one-stage models tend to have lower performance. In contrast, single-stage models have a shorter inference time, performing real-time tasks, while two-stage models have a much longer inference time \citep{objectdetection2}. In our case, the efficacy of the cell extraction task is measured by the accuracy, not by the performance speed. For this reason, a region-based architecture was preferred to accomplish this task. Hence, the architecture of the model was readjusted to meet our performance requirements, using only the \textit{region proposal functionality} and casting aside the classification part. This way, training costs and inference time are reduced.


\begin{figure*}[!ht]
 \centering
  \includegraphics[width=0.9\linewidth]{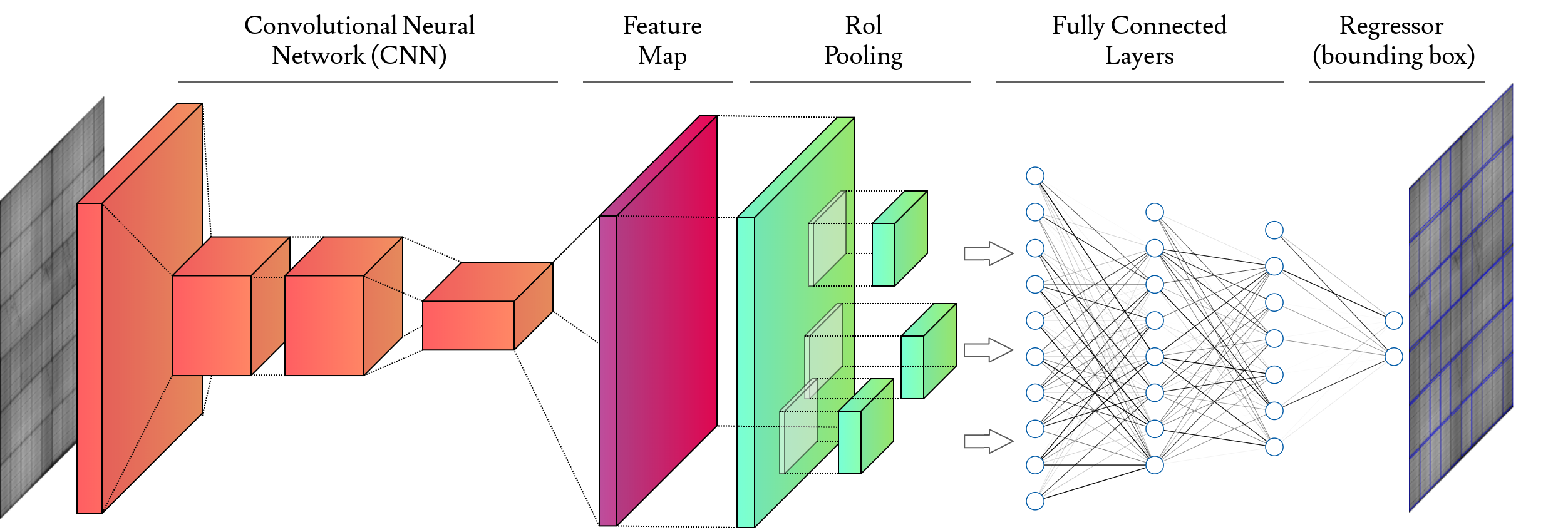}
  \caption{\textit{Modified Faster R-CNN}: Structural overview of the object detection architecture used to locate PV cells from EL panel images. The feature map is created using an convolutional neural network called ResNet-101 \citep{resnet}. Also, the fully convolutional layer has been modified to only regress the coordinates of cells, skipping classification part of the original Faster R-CNN.  }
    \label{fig:faster}
\end{figure*}
\begin{figure}[!b]
 \centering
  \includegraphics[width=\linewidth]{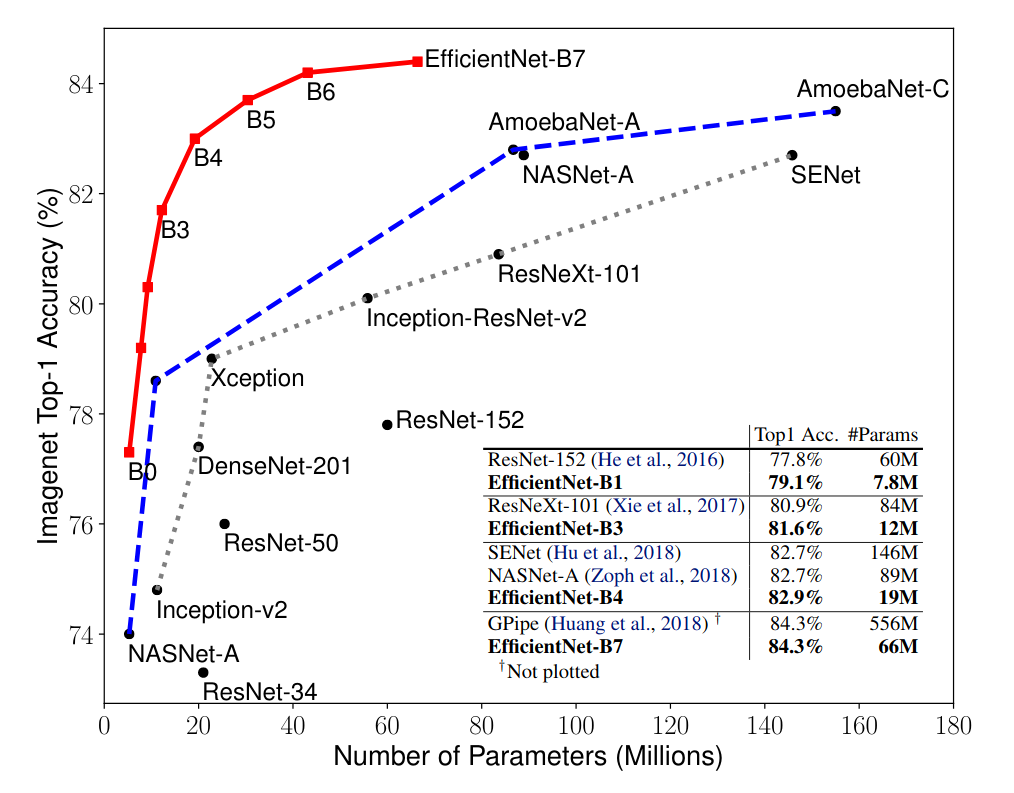}
  \caption{Comparison of ImageNet\citep{imagenet} accuracy of different models with respect to the amount of neurons in millions. The points in the red line represents the variations of \textit{EfficientNet} and the red line the state-of-the-art. \textit{source:} \citep{efficientnet}}
    \label{fig:efficientnet}
\end{figure}

Entire PV panel images are fed into the cell detection process, without any surrounding elements. Each panel contains a certain amount of adjacent cells, and all cells are expected to have the same characteristics, such as type of cell, shape, etc.

PV module images do not have the complexity of other types of images, such as the ones found on the Imagenet dataset \citep{imagenet}. A typical PV cell has a uniform color background and several horizontal or vertical lines (busbars). For the task of cell detection, the low structural complexity of the PV cells does not require a sophisticated neural network architecture to achieve good performance. \textit{Faster R-CNN} provides the perfect trade-off between performance and training computational cost for this specific task. This architecture was proposed by \cite{faster_r-cnn}, and obtained $69.9\%$ mAP an accuracy in PASCAL VOC challenge.

Faster R-CNN uses a \textit{Region Proposal Network}(RPN) based on VGG-16 \citep{vgg-16} classification CNN, to create region proposals from the input images. RPN slides \textit{k} \textit{anchor boxes} (sliding-windows) over the convolutional feature map of the image and generates all region proposals \textit{(see \cref{fig:faster})}.

We propose to modify the Faster R-CNN architecture to adjust the network to perform this task. The modification consists of the use of ResNet101 \citep{resnet} for the creation of the convolutional feature map, instead of using the VGG-16. This change was justified based on the ability of ResNet to extract features from images avoiding over-fitting, which is possible due to a large number of layers and the \textit{skip connections}.

\subsection{Classification module}
\label{section:classification_module}

The classification module aims to discriminate non-defective from defective PV cells. This module as well as the detection module can be addressed using classic computer vision methods, but, as was explained above, deep learning methods provide more accurate and robust solutions. The application of such methods requires the models to be trained using data-driven supervised techniques.

As explained above, \cite{deitsch_2019} evidenced the benefits of using deep learning techniques comparing the anomaly classification performance of a VGG-19 with an SVM. Subsequently, novel anomaly detection approaches have adopted these techniques. However, each approach uses distinct methods, datasets and performance metrics, which makes the comparison inequitable. In our case the objective of the classification module is to classify a PV cell as non-defective or defective, using as an input an EL image of a PV cell. Therefore, the approaches of \cite{tang2020deep,chen2019detection} are not comparable with the aim of this classification model.


For the classification module, a novel architecture called \textit{EfficientNet} has been selected, which was proposed by \cite{efficientnet}. This network outperformed the state-of-the-art on the classification contest ImageNet \citep{imagenet} while having $88.1\%$ fewer neurons than the previous top of the ranking.  The lightness and efficiency of this network achieve better performance with less training cost, thus it is feasible to address more complex tasks. 

\begin{figure}[!t]
\centering
\subfloat[Non-defective.]{\includegraphics[width=0.35\columnwidth]{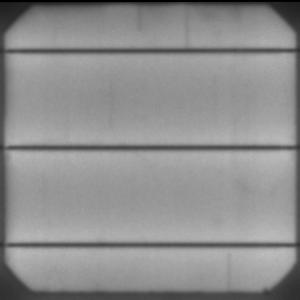}} 
\qquad
\subfloat[Defective.]{\includegraphics[width=0.35\columnwidth]{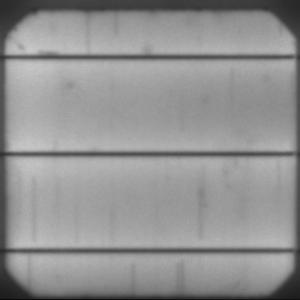}}

\caption{Representation of the strong resemblance that can exist between non-defective and defective PV cells, making it harder for a classifier to distinguish between one class and another.}
\label{fig:similar}
\end{figure}
 

The classification problem of PV cell condition is not complex in terms of domain or number of classes, but in terms of the distinctive features of each class \textit{(see \cref{fig:similar})}. Most of the anomalies that make a cell defective are micro-cracks or dents, which can be confused with the noise generated in the capture.

Therefore, a model that discerns non-defective and defective PV panels requires long training periods to be able to learn the characteristic features of each class. 
It was decided to use a smaller but more precise variation of the network, so that the model could be trained faster. The selected network is \textit{EfficientNet-B1}, which contains 7.8M neurons and achieves $79.2 \%$ accuracy on ImageNet \textit{(see \cref{fig:efficientnet})}.

When presented with a new image, these models estimate the uncertainty about the predicted class label i.e. how sure they are about the prediction. 
This uncertainty may be high on borderline cases in which the image has no clear association to neither defective nor non-defective PV modules. In pursuance of minimizing the impact of this uncertainty on the detection performance, it was decided to label as defective all the images classified by the model as non-defective with lower feasibility than 70\%. This policy generates more false positives (FP), as more non-defective images are classified as defective, resulting in a reduction in the number of false negatives (FN). However, this procedure does not affect the performance and accuracy of the entire pipeline, since FP images do not generate a damaged area segment when being evaluated by the segmentation module. In such case, the output of the segmentation model will be the original PV cell without anomaly annotation.


\subsection{Segmentation module}
\label{section:segmentation_module}

The classification module determines whether a solar cell is non-defective or defective, but it can not characterize the anomaly that makes it defective. The defect characterization requires the observation and parsing of higher-level patterns, which, in turn, implies superior annotation complexity. Therefore, the segmentation phase aims to extract from an image the area that makes the cell defective.

\paragraph{Dissuasive approaches}
\label{section:dissuasive_approaches}

The challenge of object segmentation is widely addressed in various computer vision tasks and other areas as well. Lately, with the emergence of deep learning techniques, extracting masks from objects is being undertaken with high-level models. As has been mentioned above, training supervised deep learning models generally require annotated large datasets, making data gathering a critical task.


Most domains, such as the one under discussion in this article, lack a robust and annotated dataset. Therefore, an alternative method that does not require specific annotations is required for the object segmentation task. An unsupervised learning technique could be an alternative to create a segmentation model, but the lack of high-level annotations makes the learning process tough and mostly ineffective.

\paragraph{Weakly supervised proposal}
\label{section:weakly_supervised_proposal}

Considering the limitations of both supervised and unsupervised segmentation techniques, it was decided to take advantage of lower-level annotations (classification tags), which are much easier to collect. These annotations can be used by a weakly supervised model to infer high-level information from the images.



Conversely, the classification model must distinguish between non-defective and defective PV cells, and this information can be used as lower-quality labels to train the segmentation model. Hence, the label of the cell (defective / non-defective) is used to train a model to learn the actual distribution of non-defective PV cells. This model then processes defective cell images and infers the deviation from the learned distribution. In this manner, this model is able to detect the area of the image outlying the distribution of non-defective cells.

Outlier detection can be addressed using deep learning or conventional techniques, depending on the domain's abstraction. The outliers present on images live in a $n$-dimensional space, $n$ being the number of pixels in the image. 
The presence of multivariate outliers complicates the use of conventional detection techniques but leaves the way open for more advanced and complex models, such as deep learning models. Hence, it was concluded that a deep learning model was the best approach for this task.

\paragraph{Generative models}
\label{section:generative_models}



Generative models aim to learn the distribution of original data and are able to generate new data with some variations. Among the family of generative models, it was decided to use an autoencoder since its objective is to learn a representation of data for noise reduction rather than to generate new data.
In this phase, the goal is to train an autoencoder model with non-defective cell images, learning the distribution of these images, so that the model can replicate the training set. Later, when the model encodes an image of a defective cell, due to the intrinsic similarity with the codification of non-defective cell images, it will decode the defective cell image as a non-defective version of it. This way, by subtracting the generated image from the original cell image, the damaged area is segmented.

\begin{figure*}[hbt!]
 \centering
  \includegraphics[width=\linewidth]{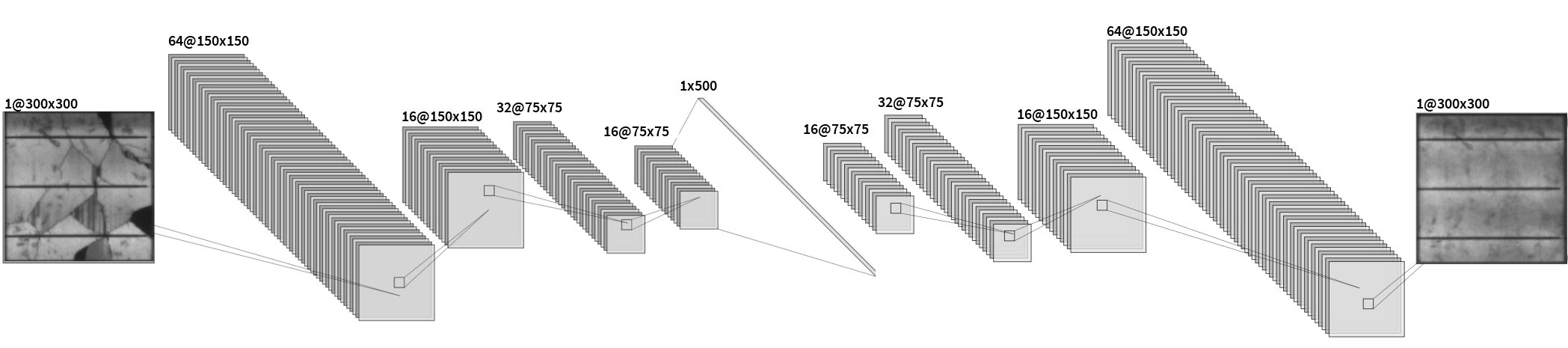}
  \caption{Visualization of the autoencoder topology for the segmentation phase.}
    \label{fig:autoencoder}
\end{figure*}

\paragraph{Autoencoder}
\label{section:autoencoder}

The autoencoder is composed of two symmetrical CNN architectures: encoder and decoder. The encoder (Enc) receives an image $I$ of shape $300 \times 300 \times 1$ and converts it to a $d$-dimension array called \textit{latent space}. The value of $d$ determines the capability of the autoencoder to learn the most complex and finer details of the data distribution. In case the value of $d$ is higher than the input data shape, the autoencoder tends to simply copy the input image. The decoder (Dec) takes in the latent space and processes it using CNNs, generating an output image $\hat{I}$ of shape $300 \times 300 \times 1$. 

\begin{equation} \label{eqn_example} 
 \hat{I} = Dec(Enc(I)) = I.
\end{equation}

Since the dataset available (described in Section 5) is composed of several types of solar cells, the autoencoder has to learn the actual distribution of each cell type. Different cell types alter the shape, the material texture and the number of busbars of the cell. 
Despite this diversity, the autoencoder should be able to generate new images without vanishing the representation of the main characteristics. To accomplish a satisfactory performance, a custom CNN architecture has been designed to convey high-level features to the latent space \textit{(see \cref{table:autoencoder})}.

\begin{table}[htbp!]
    \centering
    \resizebox{0.9\linewidth}{!}{%
    \begin{tabular}{ccccc}
        \toprule
        \textbf{Layer} & \textbf{Output Shape} & \textbf{Filters} & \textbf{Kernel} & \textbf{Stride}\\ \toprule
        Input & $300\times300\times1$ \\
        Conv2D & $150\times150\times64$& $64$ & $4\times4$ & $2\times2$ \\
        Conv2D &$150\times150\times16$& $16$ & $3\times3$ & $1\times1$ \\
        Conv2D & $75\times75\times32$ & $32$ & $4\times4$ & $2\times2$ \\
        Conv2D & $75\times75\times16$ & $16$ & $3\times3$ & $1\times1$ \\
        Flatten & $90000$ \\
        Dense & $500$ \\ 
        Dense & $90000$ \\ 
        Deconv2D & $75\times75\times16$ & $16$ & $3\times3$ & $1\times1$ \\
        Deconv2D & $75\times75\times32$ & $32$ & $4\times4$ & $2\times2$ \\
        Deconv2D & $150\times150\times16$ & $16$ & $3\times3$ & $1\times1$ \\
        Deconv2D &  $150\times150\times64$& $64$ & $4\times4$ & $2\times2$ \\
        Output & $300\times300\times1$  \\
        \bottomrule
    \end{tabular}
    }
    \caption{Autoencoder topology for the segmentation phase.}
    \label{table:autoencoder}
\end{table}

The small amount of convolutional layers allows the cell characteristics to persist during convolutions. The downsampling feature maps have only been applied a couple of times to prevent high-level features from vanishing during compression.

\paragraph{Loss function}
\label{section:loss_function}


According to the literature, the most popular loss functions are based on per-pixel error measurement, which assumes the existence of independence between the neighbors of each pixel. \cite{segmentationStructural} proved that using a loss function that captures local inter-dependencies between image regions drastically improves autoencoders' performance. Following their approach, the structural similarity index metric (SSIM) by \cite{SSIM} was selected as the loss function. This metric takes into account luminance \textit{l}, contrast \textit{c} and structure \textit{s} to compute the similarity between two images. 

\begin{equation} \label{ssim} 
SSIM(x,y)=l(x,y)^{\alpha} c(x,y)^{\beta} s(x,y)^{\gamma}.
\end{equation} 

SSIM is measured using a sliding window of $K \times K$ over the image and finally computing the average of the measurements. A low $K$-value leads to a slower and more expensive learning process but a better capability of the autoencoder to represent more precisely the real distribution of the training set. This metric \textit{(see \cref{ssim})} returns a value in the range $[-1,1]$, where $1$ means $x$ and $y$ are identical and $-1$ means that they are completely different. In our case the objective is to maximize the value of SSIM, so the loss function will be the minimization of the negative SSIM \textit{(see \cref{our_loss})}.

\begin{equation} \label{our_loss} 
Loss(\hat{I},I) = -SSIM(\hat{I},I).
\end{equation}

\paragraph{Anomaly Extraction}
\label{section:anomaly_extraction}

Once the model has learned the real distribution of the training set of non-defective photovoltaic cells, the autoencoder will process defective cells ($I$) and create their non-defective version ($\hat{I}$).

\begin{equation} \label{segmentation} 
D = SSIM(I-\hat{I}).
\end{equation} 

After the autoencoder generates the new image ($\hat{I}$), the difference between both images is computed ($D$), as expressed by the \cref{segmentation}. SSIM was used to calculate the difference, for the same reasons stated above. 
The matrix $D$ has the same dimension as the images with values between $-1$ and $1$. Each scalar value represents the SSIM of the sliding window where that pixel is located. If the value is near $1$ means that the difference in that part of the image is minimal, whereas a value close to $-1$ means there is a significant difference.
In order to reduce noise and focus exclusively on the parts where the difference between both images is notable, the image is post-processed using \textit{Otsu's} thresholding technique \citep{otsu}, which creates a \textit{binarized} image ($B$).


\begin{algorithm}
\caption{Anomaly Segmentation}
\label{alg:segmentation}
    \begin{algorithmic}[!ht]
     \renewcommand{\algorithmicrequire}{\textbf{Input:} Original Image (I)}
     \renewcommand{\algorithmicensure}{\textbf{Output:} Segmentation of the anomalies of I(S)}
     \REQUIRE 
     \ENSURE 
      \STATE {$\hat{I} = autoencoder(I)$}
      \STATE {$D = SSIM(I,\hat{I})$}
      \STATE {$B = threshold\_otsu(D)$}
      \STATE {$S = I+red(B*255)$}
     \RETURN $S$ 
     \end{algorithmic} 
\end{algorithm}

\section{Datasets}
\label{section:datasets}

Capturing electroluminescence images of photovoltaic panels is usually made when the panel is mounted at the factory, but it can also be captured at various stages of the project. This fact is due to two reasons: first, a capture of this type of image requires very specific and restrictive environmental conditions, and second, it requires a preceding procedure that is incompatible with having the panel operating at the plant. In this article, two datasets of electroluminescence images have been used to validate the proposed pipeline: a) \textit{ELPV} and b) \textit{TecnaliaPR}. In this section a detailed explanation of both datasets is presented.

\begin{figure}[!t]
 \centering
  \includegraphics[width=0.9\linewidth]{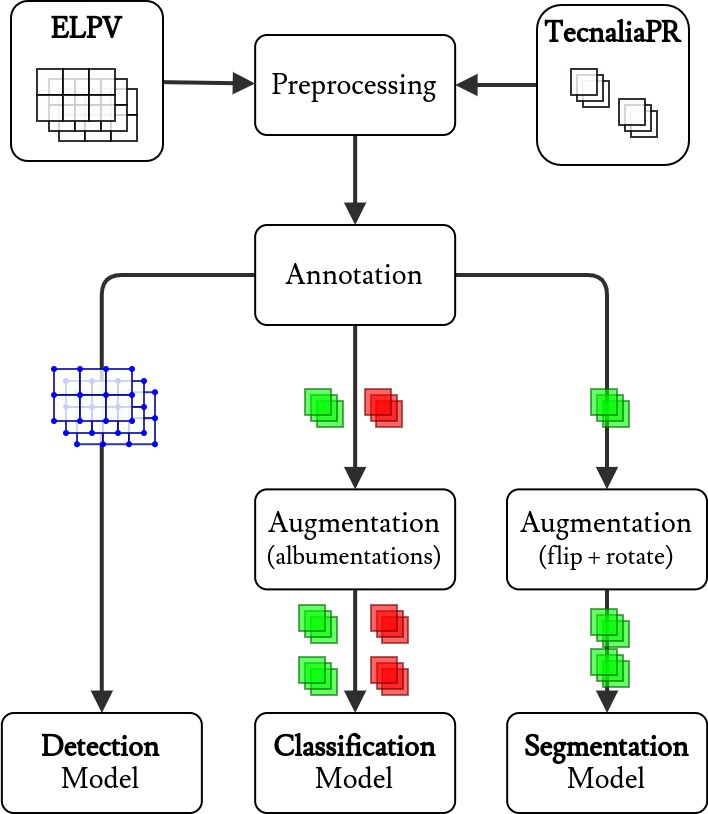}
  \caption{Data processing pipeline: preprocessing, annotation and augmentation required for each model}
  \label{fig:datasets}
 \end{figure}

\subsection{ELPV}
\label{section:elpv}

\begin{figure}[!ht]
 \centering
  \includegraphics[width=0.66\linewidth]{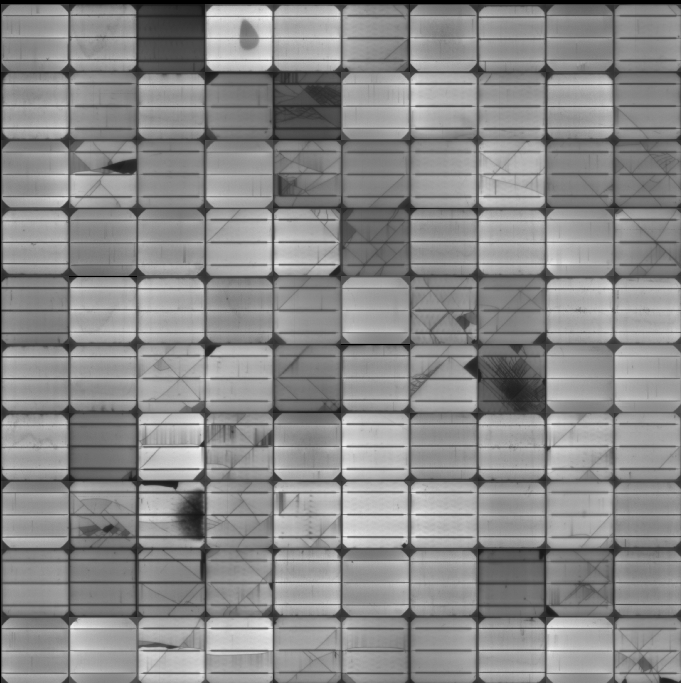}
  \caption{Sample images from the electroluminescence cell imagery dataset \citep{elpv1}. }
  \label{fig:cell}
 \end{figure}

The ELPV dataset is provided by \cite{elpv1} on their \textit{GitHub} repository.
The dataset contains $2,624$ sample images of non-defective ($1116$) and defective ($1508$) solar cells, with defects ranging from micro-cracks to completely disconnected cells and mechanically induced cracks (e.g. electrically insulated or conducting cracks, or cell cracks due to soldering) \citep{deitsch_2018}. It is important to emphasize that each image represents one sample \textit{(see \cref{fig:cell})} of a photovoltaic panel and not an entire panel. Hence, the images are small-sized, $300\times300$ pixels, and 8-bit grayscaled.
The images were obtained from 42 photovoltaic panels and were normalized concerning perspective and size. 


Considering the intrinsic difficulty of detecting the degree of degradation or damage of a photovoltaic cell, experts annotated the defect likelihood of each cell. Thus, every image is annotated with a value that represents the probability of the cell being damaged. These probabilities are discretized to have a limited number of classes (\textit{classification}), instead of having a continuous one (\textit{regression}). This discretization process leads to a four-class classification ($0.0/0.33/0.66/1.0$) that has been further relabeled into two classes: non-defective and defective. When two labels refer to similar classes and the difference between them is hard to detect, the uncertainty to train a precise \textit{deep learning} model increases exponentially. In order to avoid an increase in uncertainty, the most common practice is to unify both classes. In this case, the relabeling process has separated the cells with defects likelihood above $0.33$ as defective and the ones with defects likelihood equal to $0.0$ as non-defective \textit{(see \cref{table:elpv1})}.

It is also important to keep in mind other annotation features to explore and evaluate imagery datasets. In this case, there is only one extra feature apart from the label, which is the type of solar cell. 
Photovoltaic panels can be constructed with a wide variety of materials. Among this variety, the panels used to generate the ELPV dataset were made from \textit{Monocrystalline} and \textit{Polycrystalline} silicon. 
\cref{table:elpv1} summarizes the number of cells in the ELPV dataset, separated by cell type and by defect likelihood.

\begin{table}[htbp!]
\centering
\resizebox{0.85\linewidth}{!}{%
\begin{tabular}{cccc}
\toprule \textbf{Class}  & \begin{tabular}[c]{@{}c@{}}\textbf{Defect }\\\textbf{likelihood} \end{tabular} & \textbf{Cell Type}  & \textbf{Quantity}  \\ \toprule
\multirow{2}{*}{Non-defective} & \multirow{2}{*}{0,0} & Monocrystalline & 588 \\ 
&  & Polycrystalline & 920 \\ \midrule
\multirow{6}{*}{Defective} 
 & \multirow{2}{*}{0,33} & Monocrystalline & 117 \\
 &  & Polycrystalline & 178 \\ \cmidrule{2-4}
 & \multirow{2}{*}{0,66} & Monocrystalline & 56 \\
 &  & Polycrystalline & 50 \\ \cmidrule{2-4}
 & \multirow{2}{*}{1,00} & Monocrystalline & 313 \\
 &  & Polycrystalline & 402 \\ \bottomrule
\end{tabular}
}
\caption{ELPV dataset: number of cells per type and defect likelihood}
\label{table:elpv1}
\end{table}

Recalling that electroluminescence images are captured at the end of the fabrication process, finding defective cell images is very unusual due to the lack of defective fabrication cases. This is why the number of annotations of each class is quite unbalanced: $1116$ defective vs $1508$ non-defective.

\subsection{ELPV Augmented}
\label{section:elpv_augmented}

Applying data augmentation techniques enriches the data and makes models more robust. In the ELPV dataset, despite having a large number of images and annotations, all the data was captured from the same orientation and under similar conditions making the dataset less general and vulnerable to small variations. 
Moreover, analyzing the number of images of each type of cell \textit{(see \cref{table:elpv1})}, it can be observed that there are more \textit{Polycrystalline} cell images ($1550$) than \textit{Monocrystalline} ($1074$). A balanced dataset in terms of the variety of domain examples can be very beneficial to the deep learning model to generalize out-of-sample data. In the ELPV dataset, the type of cell is the variation on domain examples.

In this sense, data augmentation techniques can help to balance the dataset, creating more examples of concrete types of images. Applying mathematical transformations such as flip, rotate, add noise, etc. will produce more data to train the \textit{deep learning} model. 
For the data augmentation task, a fast image augmentation Python library called \textit{albumentations} \citep{albumentations} has been used. This library provides two types of transformations: (i) \textit{Pixel-level} transformations mutate color, brightness, blur, etc. whereas (ii) \textit{Spatial-level} change images shape, structure, size, etc.
In this case, taking into account the domain and how images are captured, some transformations have been discarded. Part of the augmented images cannot be related to a real situation, such as an elastic transformation of the cell. Hence, the ELPV augmented dataset was created combining and applying the transformations included in \cref{table:transformations} to the original data.

\begin{table}[!ht]
\centering
\resizebox{0.7\linewidth}{!}{%
\begin{tabular}{cc} \toprule
 \textbf{Pixel-level}  & \textbf{Spatial-level}  \\ \toprule
Random Contrast & IAA fliplr \\
Random Gamma & IAA flipud \\
Random Brightness & IAA perspective \\
Blur & Rotate \\
Jpeg compression & Grid distortion \\
Solarize & Transpose \\
Equalize & IAA sharpen \\
ISO Noise & Optical distortion \\
Random Shadow & Horizontal flip \\
Contrast Limited AHE & Vertical flip \\ \bottomrule
\end{tabular}
}
\caption{Transformations used on data augmenting.}
\label{table:transformations}
\end{table}


The result is a dataset with $3000$ images per both \textit{non-defective} and \textit{defective} cells \textit{(see \cref{table:elpv2})}. It is important to emphasize the relevance of the relabeling, which has allowed to have less unbalanced classes for posterior balancing. If the difference in annotations between classes had been bigger (\textit{e.g. 106 images of class 0.66 and 715 of class 1.0}) the augmentation would have been applied several times over few images, resulting in very similar data.

Besides balancing the dataset regarding the number of images per class, the dataset was also balanced with respect to the number of images per type of cells.
After the augmentation process, this difference has completely vanished, as can be seen in  \cref{table:elpv2}. In the new dataset, the number of \textit{monocrystalline} cell images is $3000$ and \textit{polycrystalline} is also $3000$.
The number of images has increased from the $2624$ initial ones to the augmented $6000$, equally divided into non-defective and defective cell images. In summary, the original dataset has been modified on behalf of obtaining a more robust, balanced and complete dataset.

\begin{table}[!t]
\centering
\resizebox{0.85\linewidth}{!}{%
\begin{tabular}{cccc}
\toprule \textbf{Class}  & \textbf{Type}  & \textbf{Original}  & \textbf{Processed}~ \\ \toprule
\multicolumn{1}{l}{} & \multicolumn{1}{l}{} & \multicolumn{1}{l}{} & \multicolumn{1}{l}{} \\ \cmidrule{2-4}
\multirow{2}{*}{Non-defective} & Monocrystalline & 588 & 1500 \\
 & Polycrystalline & 920 & 1500 \\ \cmidrule[\heavyrulewidth]{2-4}
 & Total & \textbf{1508}  & \textbf{3000}  \\
\multicolumn{1}{l}{} & \multicolumn{1}{l}{} & \multicolumn{1}{l}{} & \multicolumn{1}{l}{} \\ \cmidrule{2-4}
\multirow{2}{*}{Defective} & Monocrystalline & 486 & 1500 \\
 & Polycrystalline & 630 & 1500 \\ \cmidrule[\heavyrulewidth]{2-4}
 & Total & \textbf{1116}  & \textbf{3000}  \\
\multicolumn{1}{l}{} & \multicolumn{1}{l}{} & \multicolumn{1}{l}{} & \multicolumn{1}{l}{} \\ \midrule
Total &  & \textbf{2624}  & \textbf{6000}  \\ \bottomrule
\end{tabular}
}
\caption{ELPV dataset: number of original and processed cells per type and class}
\label{table:elpv2}
\end{table}

\subsection{TecnaliaPR}
\label{section:TecnaliaPR}

The dataset is composed of 67 solar modules of different cell technologies, namely monocrystalline of 5 busbars, monocrystalline half-cells of 5 busbars and heterojunction of 3 busbars). The EL images were acquired at Tecnalia's facilities, in the framework of the PROMISE project (KK2019/00088), using an Endeas QuickSun 600Lab solar simulator with an integrated EL system. This laboratory equipment features an EL image acquisition system with 200 um resolution by means of 4 infrared CCD cameras of 8.3 MP. Each camera focuses on one-fourth of the PV module and the final image is obtained by stitching the images captured from the 4 cameras (QuickSun software). It is important to emphasize that each image represents the entire solar module and not one of the many cells (see fig. 7) of a photovoltaic module. These modules had been operating under real conditions in the outdoors of the Tecnalia's facilities and were temporarily removed to capture the images. The images are quite large, $7942 \times 4096$ pixels (later subsampled to $2111 \times 1261$), and 8-bit gray-scaled. The number of cells per module varies between 60 and 120, yielding a total of 5592 cells. Among these, $4885$ are identified as non-defective and the remaining $707$ as defective. This information has been illustrated in  \cref{table:TecnaliaPR1}. The cell images were labeled by a group of experts in defective solar cells, having various defects such as electrically insulated cracks, micro-cracks, or dark areas.

To train the cell detection model, a subset of the dataset was annotated with the bounding box of each cell. This task was carried out with \textit{LabelImg}, an image annotation tool that helps to label object bounding boxes in images \citep{tzutalin2015labelimg}.
Moreover, analyzing the number of images of each type of cell, it can be observed that there are more cell images with \textit{3 busbars} ($2592$) than with \textit{5 busbars} ($1800$) and \textit{elongated} ($1200$). Hence, this dataset is also quite unbalanced with respect to the cell class and the cell type, as was the ELPV original dataset. For that reason, the same augmentation techniques have been applied to the TecnaliaPR dataset, which is explained in the next section.

\begin{table}[t]
\centering
\resizebox{\linewidth}{!}{%
\begin{tabular}{cccccc}
\toprule \multirow{3}{*}{\textbf{Cell Type} } & \multirow{3}{*}{\textbf{Modules} } & \multirow{3}{*}{\begin{tabular}[c]{@{}c@{}}\textbf{Cells per}\\\textbf{module} \end{tabular}} & \multirow{3}{*}{\textbf{Cell Size}} & \multicolumn{2}{c}{\textbf{Cells }} \\ \cmidrule{5-6}
 &  &  & & Total & \begin{tabular}[c]{@{}c@{}}non-defective /\\defective \end{tabular} \\ \toprule
Elongated & 10 & 120 & $397 \times 682$ & 1200 & 984 / 216 \\
3 Busbars & 27 & 96 & $661 \times 512$ & 2592 & 2281 / 311 \\
5 Busbars & 30 & 60 & $794 \times 682$ & 1800 & 1620 / 180 \\ \midrule
\textbf{Total}  & \textbf{67}  & \textbf{-}  & \textbf{-}  & \textbf{5592}  & \textbf{4885 / 707} \\ \bottomrule
\end{tabular}
}
\caption{TecnaliaPR dataset: number of modules and cells per type and class}
\label{table:TecnaliaPR1}
\end{table}

\subsection{TecnaliaPR Augmented}
\label{section:TecnaliaPR_augmented}

Data augmentation techniques can help to balance the dataset, creating more examples of concrete types of images. The TecnaliaPR dataset has been augmented using the same library as with the ELPV dataset (\textit{albumentations} \citep{albumentations}). Rather than trying to generate a larger dataset, the augmentation task tries to balance the cell class and the cell type across the dataset. In this sense, the limiting category is the defective class, with $707$ images compared to the non-defective class $4885$, an almost 7 to 1 ratio. \cref{table:TecnaliaPR2} summarizes the resulting quantities of cell images from the augmentation process. The number of defective images increased to $2083$, with a balanced distribution across cell type (around $700$ images per type). The number of non-defective images, on the contrary, was filtered to $2984$. It should be noted that only $1000$ images of 3-busbar and 5-busbar were used to balance the dataset. Non-defective cells can be characterized with fewer images than defective cells. This is because defective images can have a variety of anomalies with different shapes, forms and sizes, whereas non-defective ones are more similar between them. Thus $1000$ images per class were sufficient to learn the distribution of non-defective PV cells. With this, the modified dataset was finally balanced, both regarding the cell type and the cell class.

\begin{table}[htbp!]
\centering
\resizebox{0.8\linewidth}{!}{%
\begin{tabular}{cccc}
\toprule \textbf{Class}  & \textbf{Cell Type}  & \textbf{Original}  & \textbf{Processed}  \\ \toprule
\multicolumn{1}{l}{} & \multicolumn{1}{l}{} & \multicolumn{1}{l}{} & \multicolumn{1}{l}{} \\ \cmidrule{2-4}
\multirow{3}{*}{Non-defective} & Elongated & 984 & 984 \\
 & 3 Busbars & 2281 & 1000 \\
 & 5 Busbars & 1620 & 1000 \\ \cmidrule{2-4}
 & Total & \textbf{4885}  & \textbf{2984}  \\
 &  &  &  \\ \cmidrule{2-4}
\multirow{3}{*}{Defective} & Elongated & 216 & 694 \\
 & 3 Busbars & 311 & 708 \\
 & 5 Busbars & 180 & 681 \\ \cmidrule{2-4}
 & Total & \textbf{707}  & \textbf{2083}  \\
 &  &  &  \\ \midrule
Total &  & \textbf{5592}  & \textbf{5067}  \\ \bottomrule
\end{tabular}
}
\caption{TecnaliaPR dataset: original and processed cells per type and class}
\label{table:TecnaliaPR2}
\end{table}

\section{Experimental Analysis and Results}
\label{section:experimental_analysis_and_results}

Having described the deep-learning architecture of each module, in this section the training results and the performance of the obtained models are presented. The training was performed using a GPU Tesla T4 with 16Gb of memory. It should be pointed out that even though an iterative procedure has been followed for the training of each model, in this section only the final version of the models is presented.

\subsection{Detection model}
\label{section:detection_model}

\begin{table}
\centering
\resizebox{\linewidth}{!}{%
\begin{tabular}{cccccc}
\toprule 
\multirow{2}{*}{\textbf{Cell Type} } & \multirow{2}{*}{\begin{tabular}[c]{@{}c@{}}\textbf{Cells }\\\textbf{per panel} \end{tabular}} & \multicolumn{2}{c}{\textbf{Panels } } & \multicolumn{2}{c}{\textbf{Cells } } \\ \cmidrule{3-6}
 &  & Total & Train / Val & Total & Train / Val \\ \toprule
Elongated & 120 & 10 & 8 / 2 & 1200 & 960 / 240 \\
3 Busbars & 96 & 27 & 22 / 5 & 2592 & 2112 / 480 \\
5 Busbars & 60 & 30 & 24 / 6 & 1800 & 1440 / 360 \\ \bottomrule
\end{tabular}
}
\caption{Detection Dataset: TecnaliaPR}
\label{table:detection2}
\end{table}

Training the modified Faster R-CNN model requires well-annotated images of photovoltaic panels. Each cell of the panel will have to be annotated by a bounding box containing $x_{min}, y_{min}, x_{max}, y_{max}$ coordinates and the class, which in our case is always a unique class (cell). As it has been reviewed in the previous \cref{section:datasets}, the dataset for the detection task is the expert-labeled TecnaliaPR dataset. The dataset contains 5592 annotations of cells over $67$ panel images of $2111 \times 1261 \times 1$ and has been split in two parts, training ($80 \%$) and validation ($20 \%$) sets \textit{(see \cref{table:detection2})}. 
The model was trained for $6,500$ steps, using a learning rate of $0.00003$. Every 200 steps the real performance of the model was validated using the validation set. 

Considering that in a PV module image there may be between $50$ and $100$ PV cells, the maximum of generated region proposals was set to $300$, with an objectness loss weight of $1.0$ and a classification loss weight of $0.0$. Finally, for the selection of the most promising regions the \textit{non-max suppression} (NMS) \citep{nms} threshold is set to $0.9$, to avoid the overlap between generated region proposals.


\begin{figure}[!ht]
 \centering
  \includegraphics[width=0.9\linewidth]{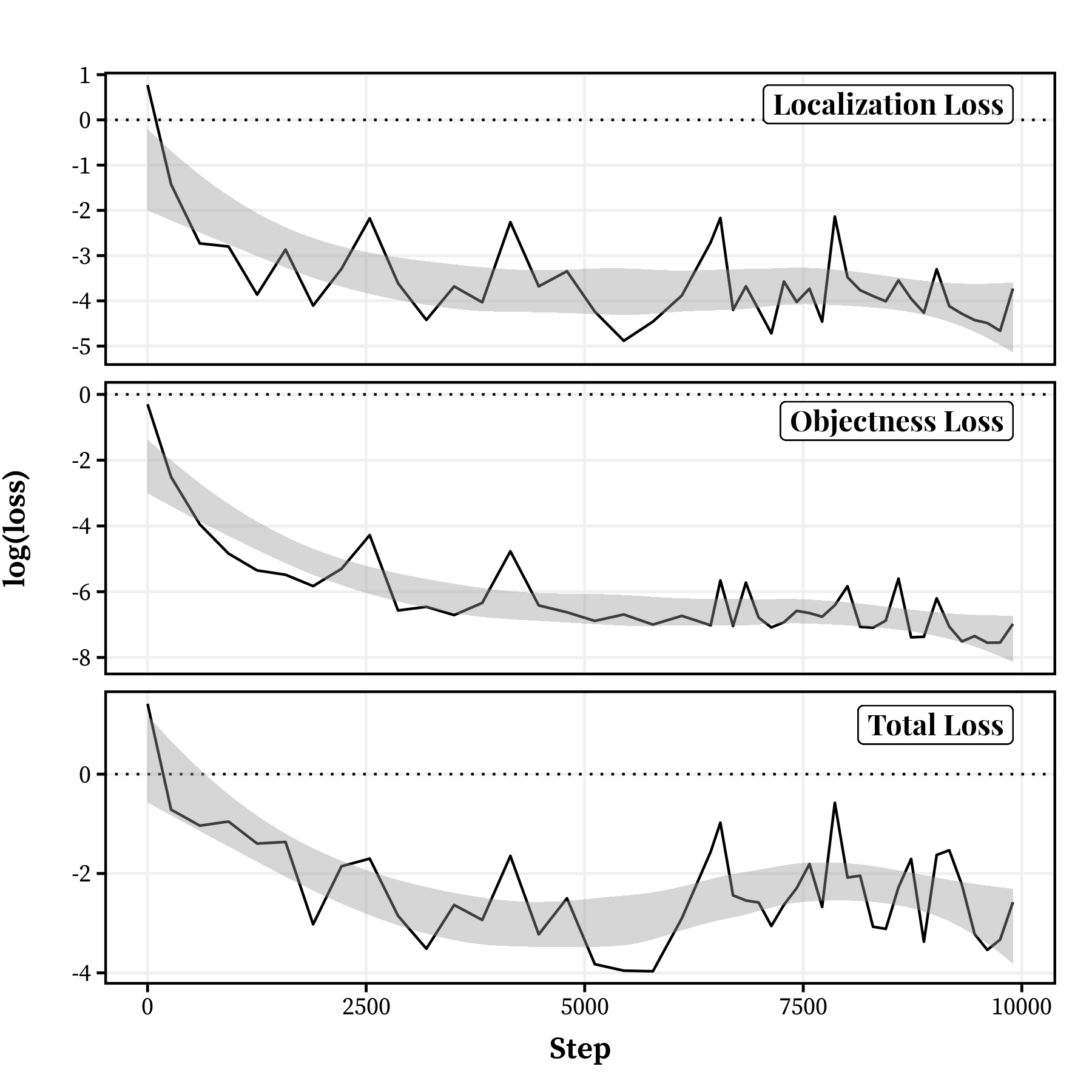}
  \caption{Detection loss in logarithmic scale, from top to bottom: localization, objectness and total loss (\textit{ETC: 40 minutes}). }
  \label{fig:detection_loss}
 \end{figure}
 
As it can be seen in the training curve \textit{(see \cref{fig:detection_loss})}, during the first steps the loss reduction was very accentuated. This usually happens when transfer learning techniques are not used. Due to this lack of previous knowledge, in the first $200$ steps the loss was high (around $4$) and once the model learned the complexity of the domain, the validation loss converged rapidly.

\begin{figure}[!htpb]
\centering
\subfloat{\includegraphics[width=0.45\columnwidth]{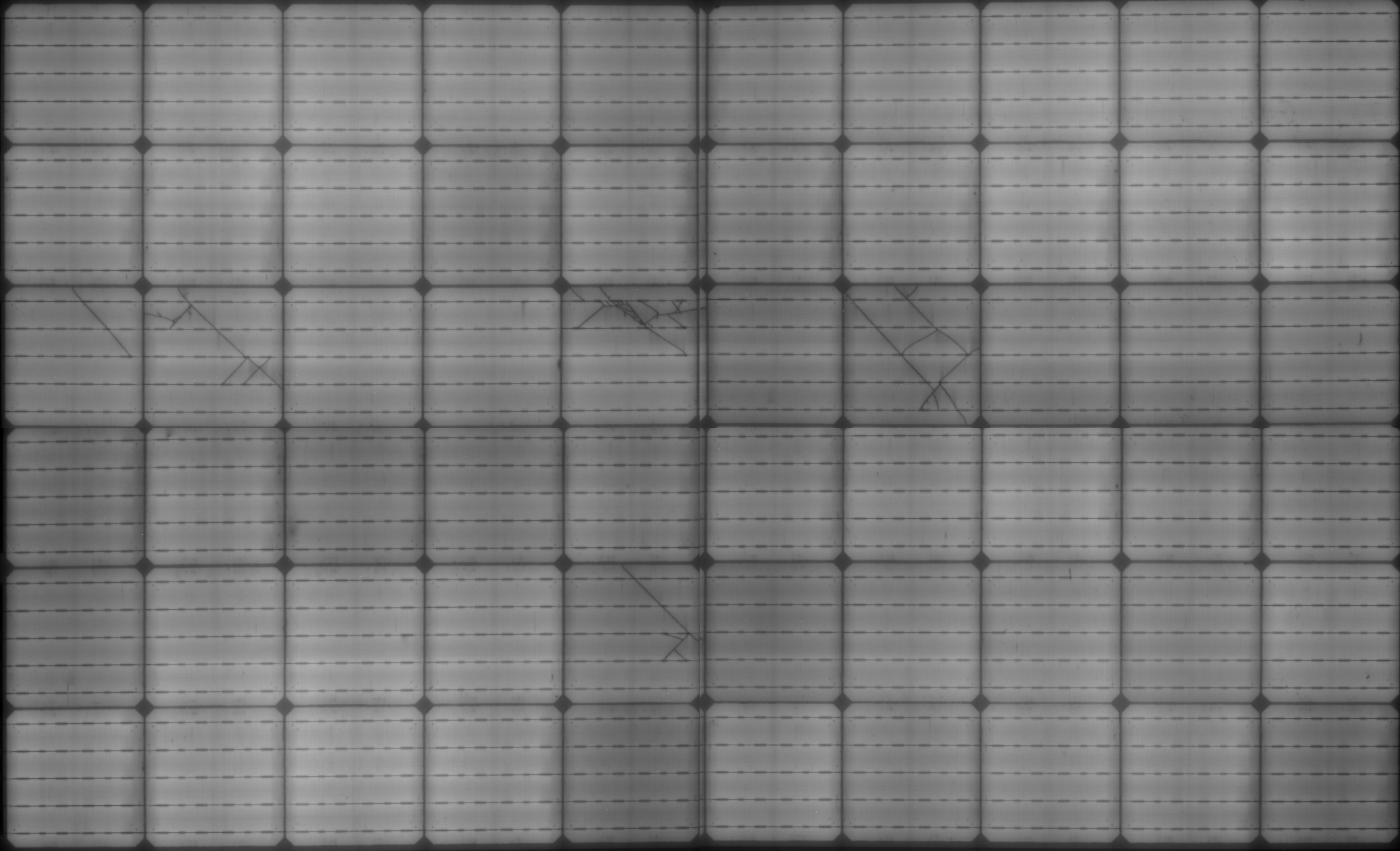}}
\qquad
\subfloat{\includegraphics[width=0.45\columnwidth]{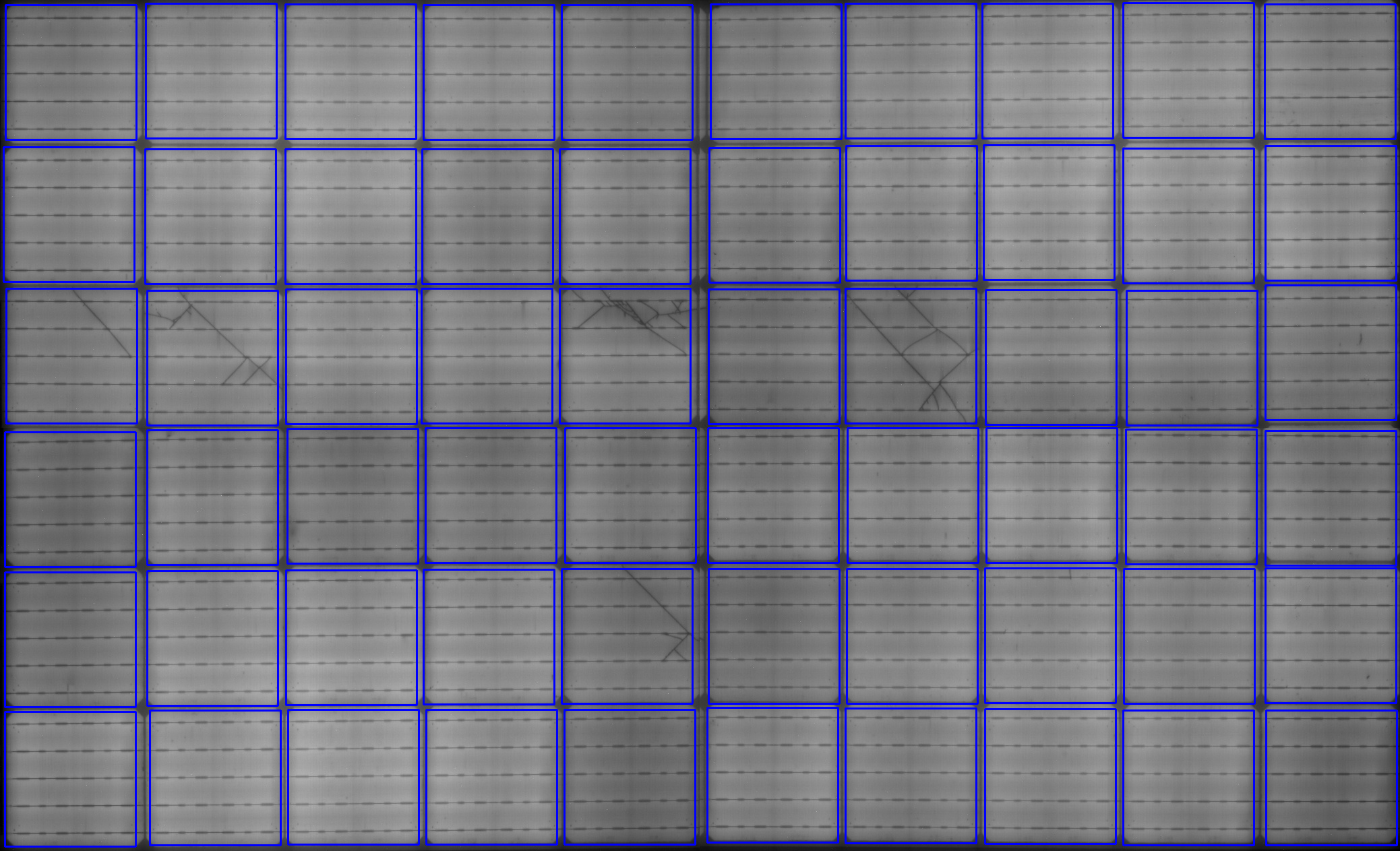}}\hfill
\subfloat{\includegraphics[width=0.45\columnwidth]{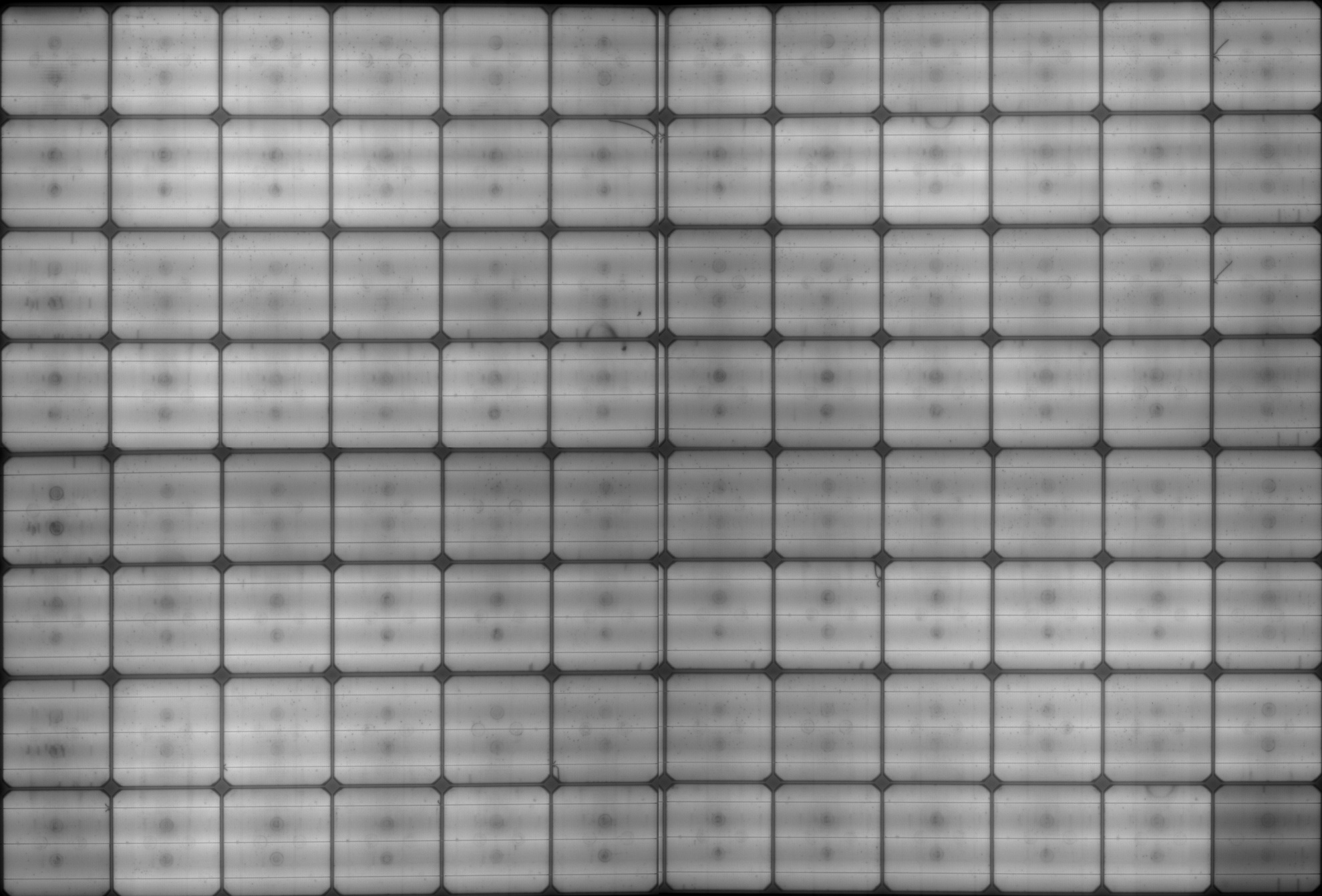}}   
\qquad
\subfloat{\includegraphics[width=0.45\columnwidth]{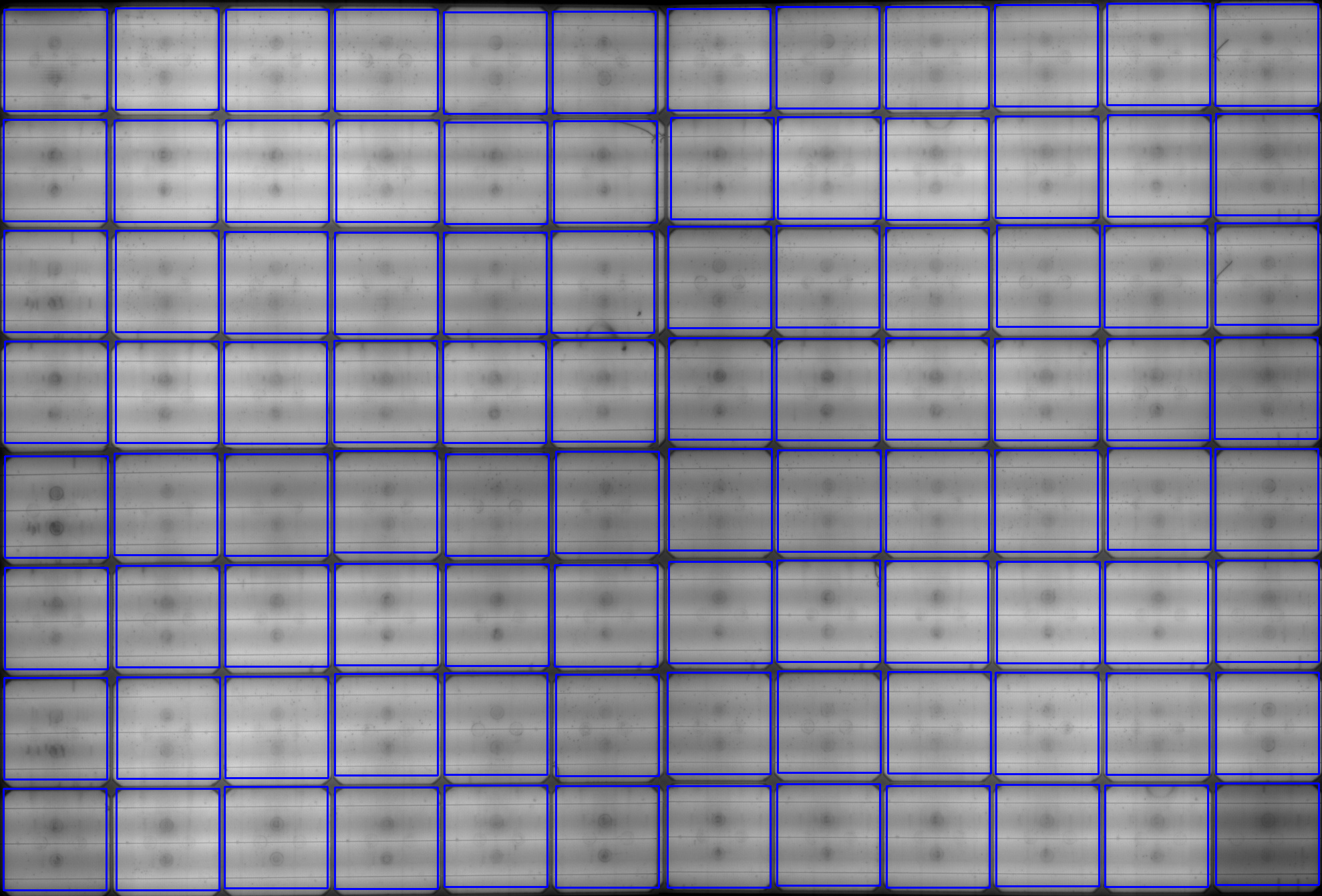}}\hfill
\subfloat{\includegraphics[width=0.45\columnwidth]{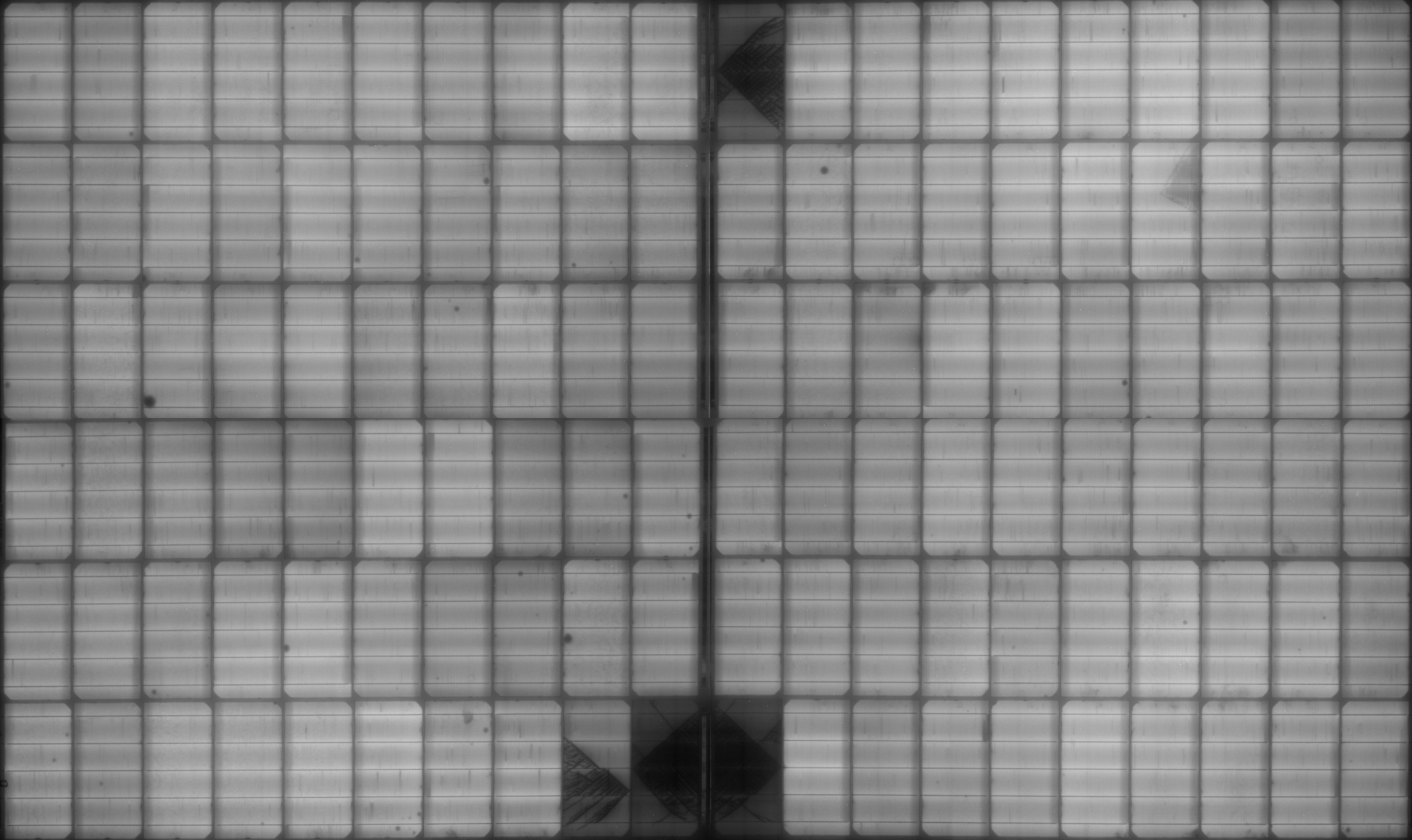}}  
\qquad
\subfloat{\includegraphics[width=0.45\columnwidth]{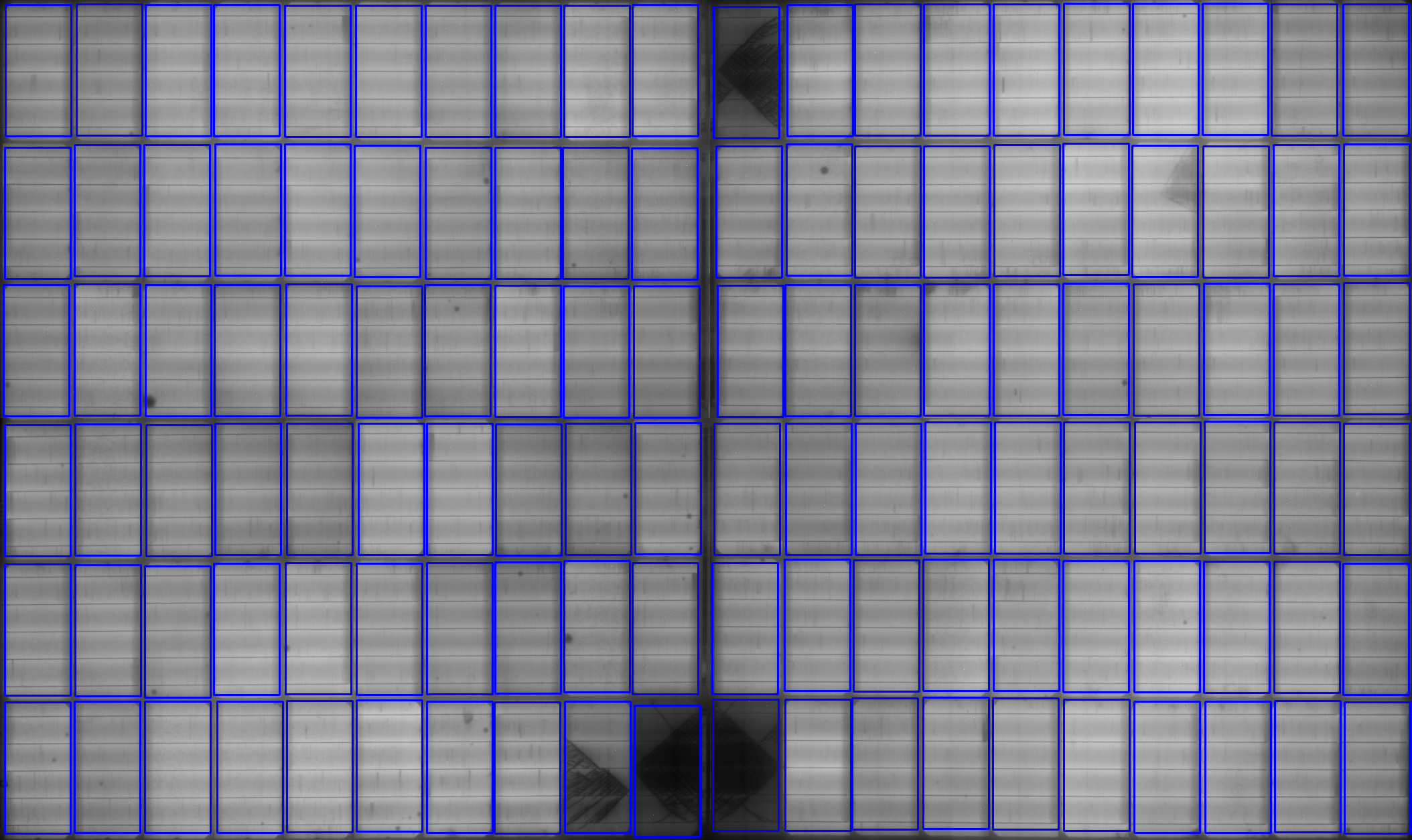}}

\caption{Representation of the performance of the Fast R-CNN detector on four solar panels, each one with different types of cells. The left side contains the original images and the right side the result of the processing done by our model, representing each detected cell in a blue bounding box.}
\label{fig:example_detection}
\end{figure}

This training process has resulted in a model with a validation loss of $0.03844$ which is a strong indicator of good performance. After the training process, the validation dataset was also used to certify the correct performance of the model and its proper functioning. The metric used to measure the model's performance was the average precision (\textit{AP)}, which is based on the object detection challenge COCO \citep{coco}. According to this measurement the cell detectors AP is $0.9936$ \textit{(see \cref{fig:example_detection})}.

\subsection{Classification model}
\label{section:classification_model}

\begin{table}[b!]
\centering
\resizebox{0.9\linewidth}{!}{%
\begin{tabular}{cccc}
\toprule
\textbf{Dataset} & \textbf{Type} & \textbf{Augmented} & \textbf{Quantity} \\ \toprule
\multirow{2}{*}{ELPV} & non-defective & Yes & 3000 \\
 & defective & Yes & 3000 \\ \midrule
\multirow{2}{*}{TecnaliaPR} & non-defective & No & 3000 \\
 & defective & Yes & 2083 \\ \toprule
 &  & \textbf{Total} & \textbf{11083} \\ 
 &  & Train/Validation & 9083/2000
\end{tabular}
}
\caption{Classification Dataset}
\label{table:classification}
\end{table}

The aim of training the \textit{EfficientNet-B1} classifier is to learn to discriminate between non-defective and defective cells, which as mentioned above, is a non-trivial task \textit{(see \cref{fig:similar})}. To accomplish such a task, it was necessary to collect as much data as possible, hence both datasets (ELPV and TecnaliaPR, explained in Sections \ref{section:elpv} and \ref{section:TecnaliaPR}) were unified. In this way, a more robust dataset was obtained, with more diversity in terms of anomalies and types of cells. The dataset for the classification model contains $11083$ images, balanced between non-defective ($6000$) and defective ($5083$) cell images, which has been split into two parts, training ($82 \%$) and validation ($18 \%$) sets. Regarding the distribution of images, there were enough defective images from the original TecnaliaPR dataset ($2083$), so it was decided not to take more from this set in order to not unbalance the training dataset \textit{(see \cref{table:classification})}.

The model has been trained for $75,000$ steps, performing a validation process every $10,000$ steps. The learning rate was set to $0.00012$ with a proportional decay every $200$ steps. The training and validation batch size was $16$ and for the optimization of the training in GPU the channel first format was used.

\begin{figure}[!t]
 \centering
  \includegraphics[width=0.9\linewidth]{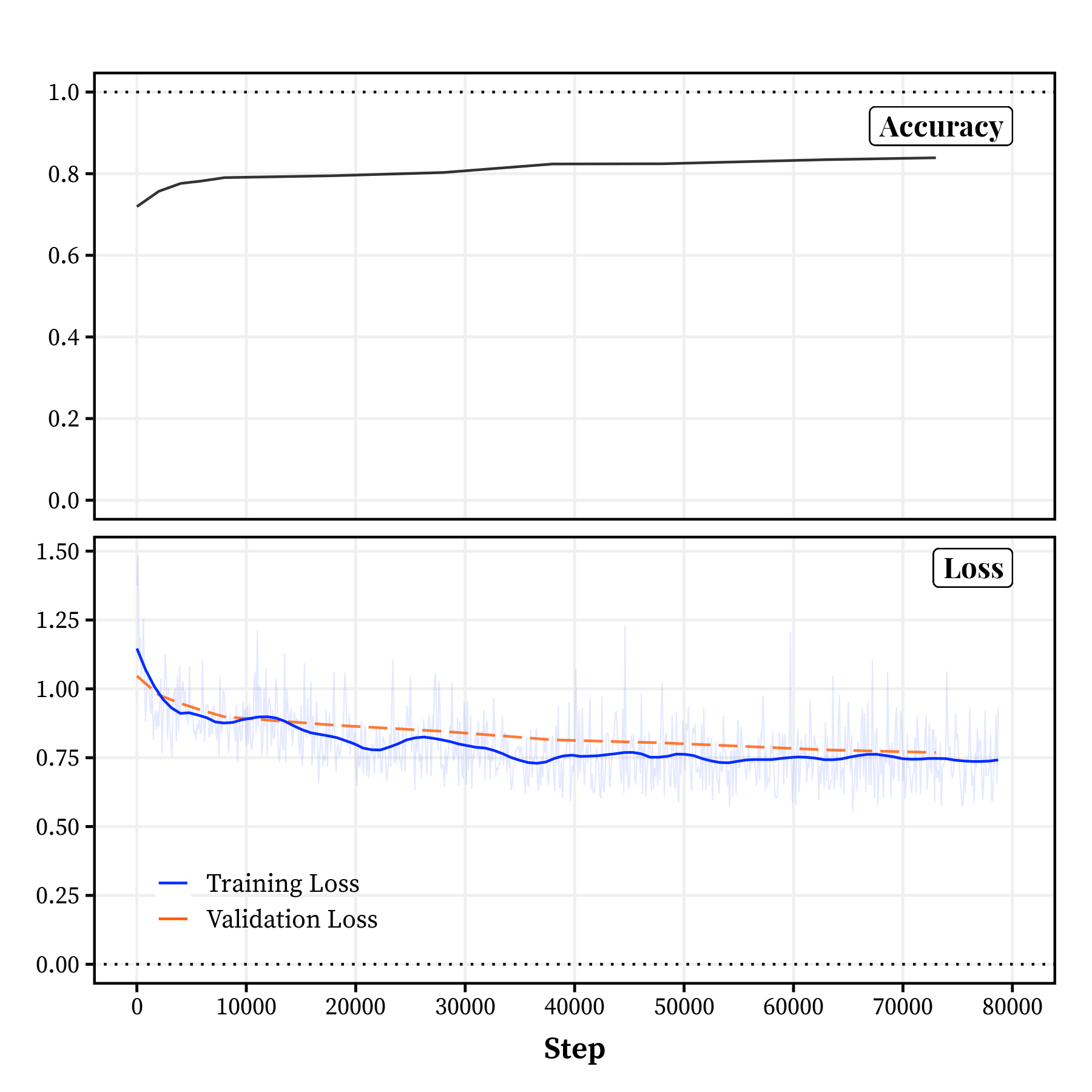}
  \caption{Classification training: accuracy and training/validation loss at each step (\textit{ETC: 960 minutes})}
  \label{fig:classification_loss}
 \end{figure}
 
Observing the evolution of the loss function during training \textit{(see \cref{fig:classification_loss})} it can be seen that the value of the loss jumps at each training step. This happens because of the great diversity of the dataset in terms of material, cell type or shape. Despite this fact, the model has gradually learned to reach a considerably low loss value for the complexity of the domain. As it can be observed, the training loss has been consistently lower than the validation loss. The validation loss has been decreasing gradually and the training loss started to converge into $0.75$ in the last epochs.
Although this may indicate that continuing the training may improve the model, it was decided to stop the training process due to its asymptotic behavior. 

\begin{figure}[!t]
\centering
\subfloat{\includegraphics[width=0.6\columnwidth]{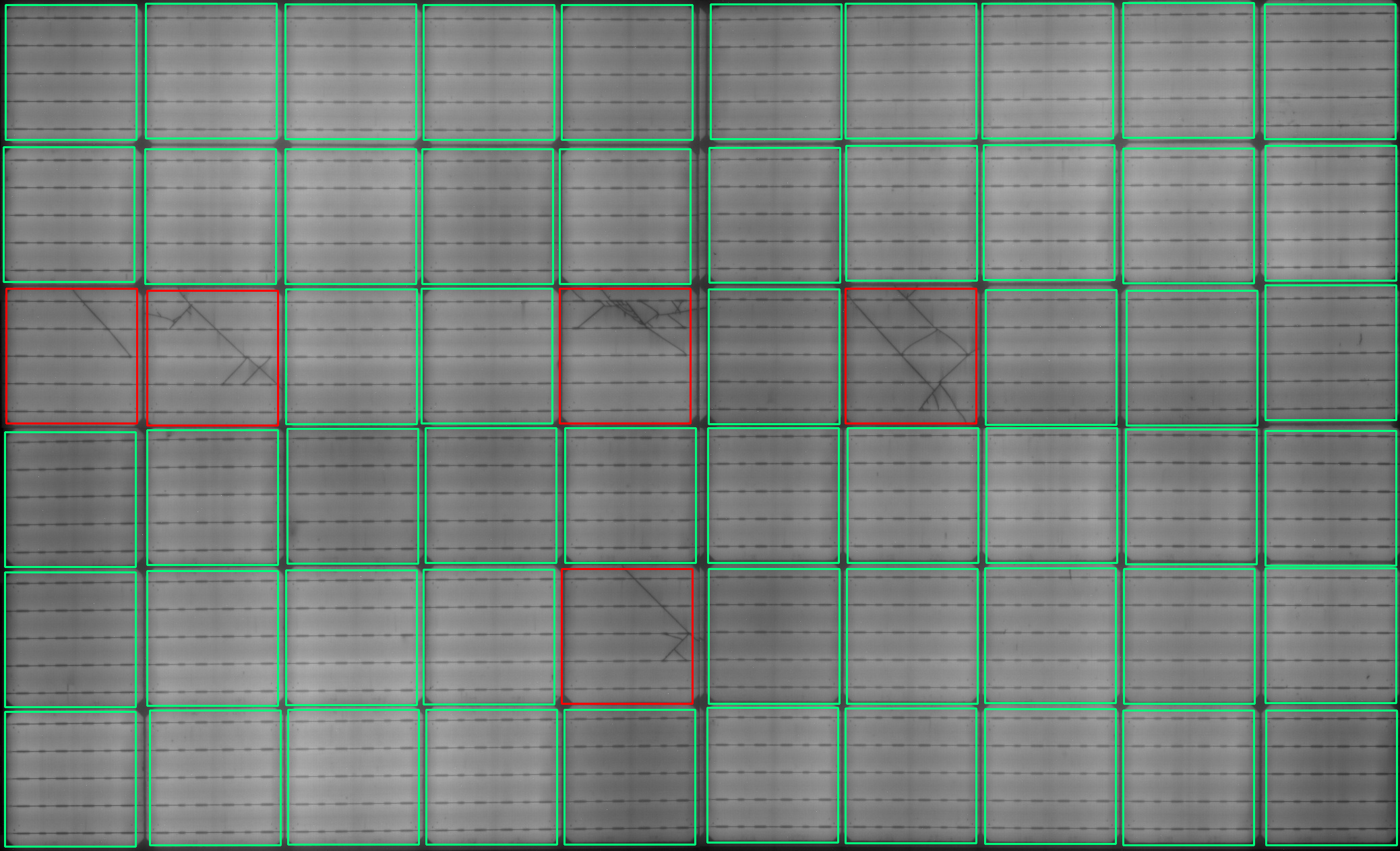}}\hfill
\subfloat{\includegraphics[width=0.6\columnwidth]{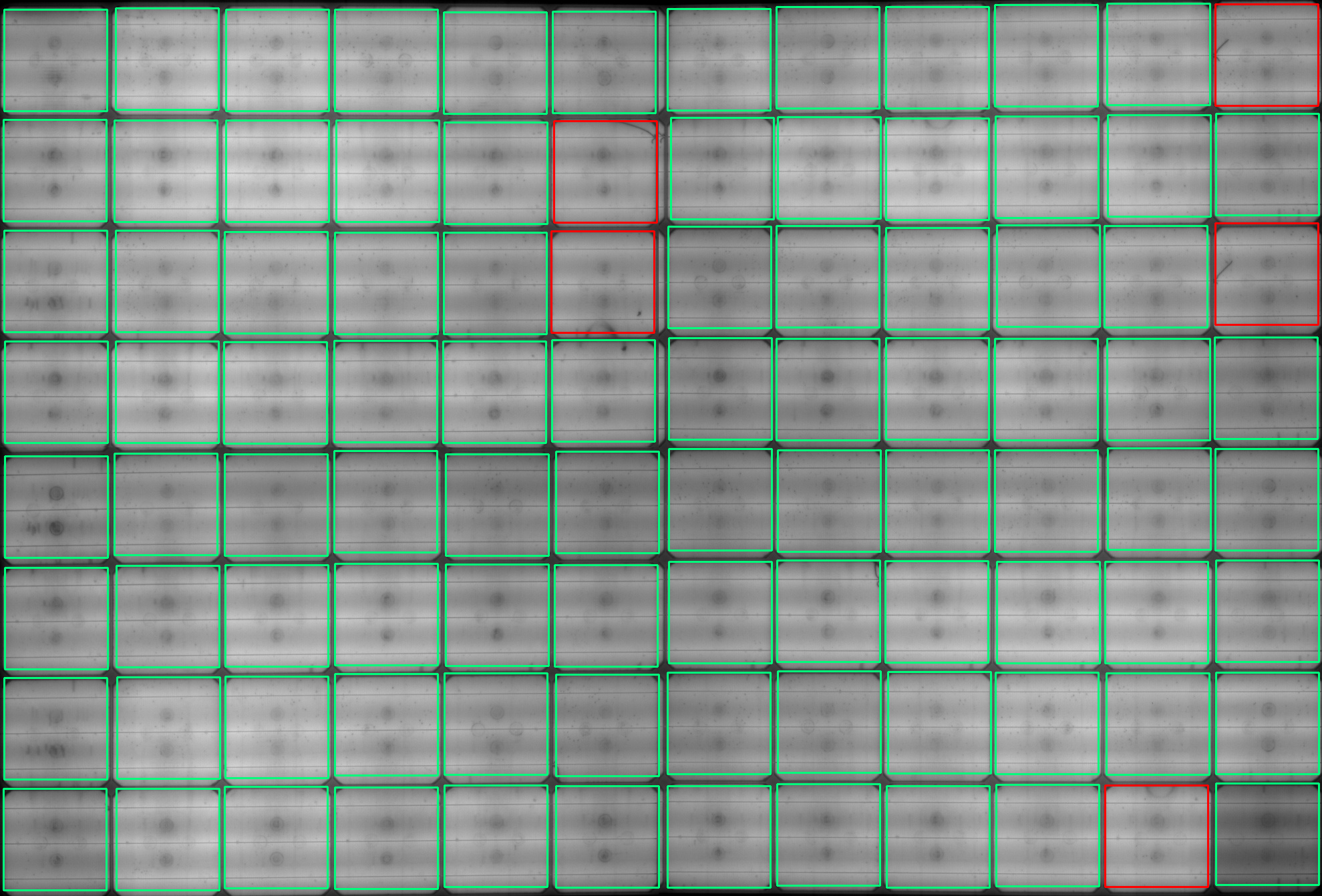}}\hfill
\subfloat{\includegraphics[width=0.6\columnwidth]{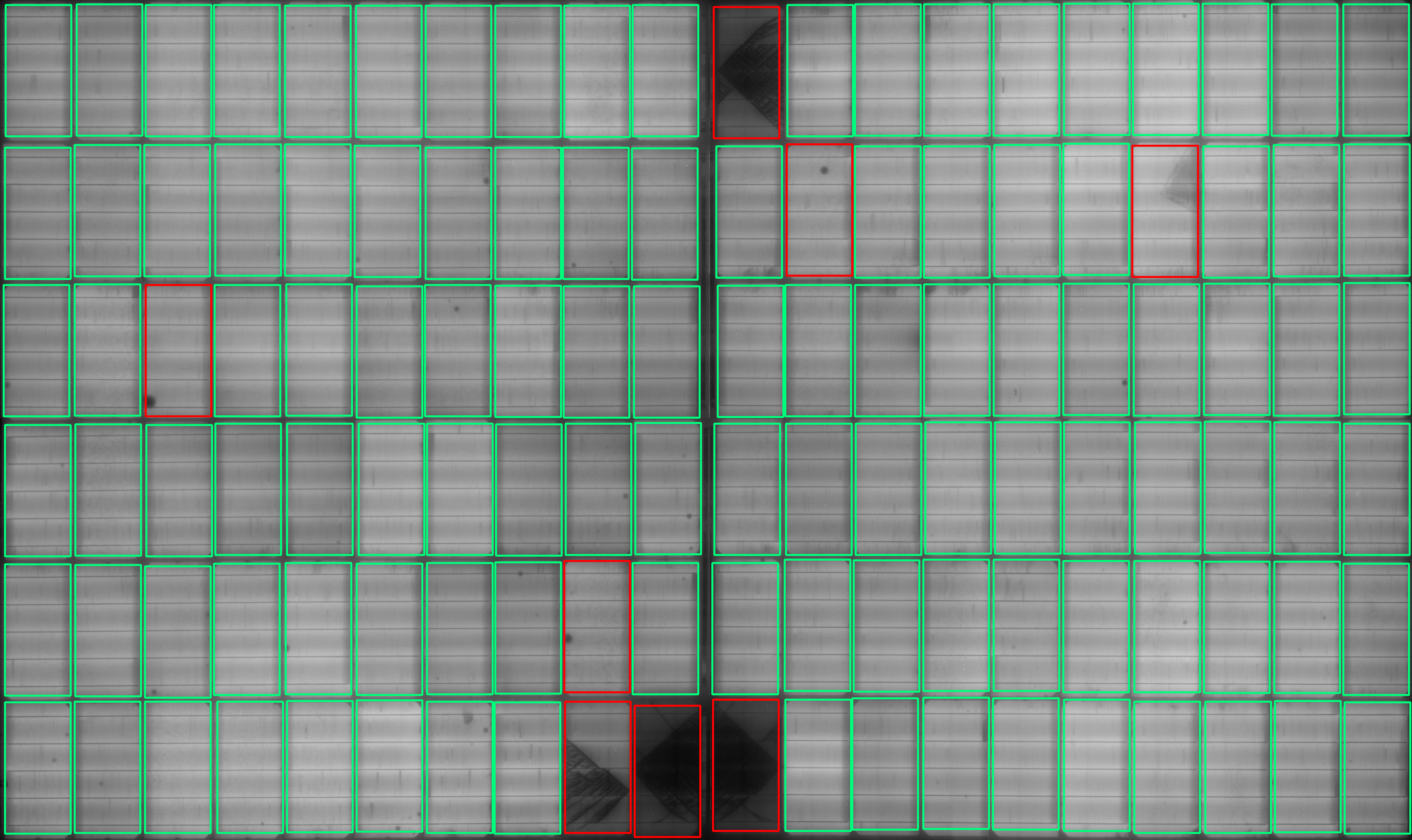}} 
\caption{Representation of the performance of the \textit{EfficientNet-B1} classifier on four solar panels, each one with different types of cells. The green boxes represent non-defective cells and the red ones defective cells. It should be noted that, in the second image, the cell in the right bottom corner was incorrectly classified by the model as non-defective. The rest of the cells were correctly classified by the model.}
\label{fig:example_classification}
\end{figure}

\begin{table}[!t]
\centering
\resizebox{0.7\linewidth}{!}{%
\begin{tabular}{cc|cc|c}
\multicolumn{2}{c}{} & \multicolumn{2}{c}{\textbf{Labels}} &  \\
 & \multicolumn{1}{l|}{} & P & N & Total \\ \cline{2-5}
\multirow{2}{*}{\textbf{Prediction}} & P & 804 & 291 & 1095 \\
 & N & 33 & 872 & 905 \\ \cline{2-5}
 & Total & 837 & 1163 & 2000
\end{tabular}
}
\caption{Confusion matrix for the classification model in the validation set}
\label{table:confusion}
\end{table}

After the training process, the performance of the model was measured, obtaining an accuracy of $0.84$. The presented model is almost as accurate as the model proposed by \cite{deitsch_2019} ($0.863$) and lags behind the precision obtained by \cite{akram2019cnn} ($0.9302$). Nevertheless, it must be stated that the complexity of the problem tackled here is higher (more cell types and diversity of anomalies), which indicates that the model is sufficiently robust and reliable to use in our pipeline \textit{(see \cref{fig:example_classification})}.

\subsection{Segmentation model}
\label{section:segmentation_model}

The training set for the segmentation model is based on original non-defective cell images from both the ELPV ($1508$) and TecnaliaPR ($4885$) datasets, as illustrated in \cref{table:segmentation}. An augmentation process has been applied to these non-defective images, going from $6393$ to a total of $11414$, based on two reasons. 

\begin{table}[!t]
\centering
\resizebox{0.8\linewidth}{!}{%
\begin{tabular}{cccc}
\toprule
\textbf{Dataset} & \textbf{Type} & \textbf{Original} & \textbf{Augmented} \\ \toprule
ELPV & non-defective & 1508 & 2692 \\
TecnaliaPR & non-defective & 4885 & 8722 \\ \toprule
Total &  & 6393 & 11414 \\ \bottomrule
\end{tabular}
}
\caption{Segmentation Dataset}
\label{table:segmentation}
\end{table}

First, it increases the amount of available data, which is beneficial for the deep learning model. Second, it balances the dataset with respect to the cell types. In fact, there are 5 main types of cells: monocrystalline and polycrystalline on ELPV; and elongated, 3 busbars, and 5 busbars on TecnaliaPR. After the augmentation, approximately 2500 images have been obtained for each cell type. The augmentation process is carried out with simple transformations, such as flipping and mirroring. More complicated transformations could introduce a considerable amount of noise, making the autoencoder less accurate.

As the training process for segmentation has a higher complexity than the detection and classification model, it implied a harder tuning. Due to the small amount of data and variety of cell types, the risk of overfitting and its consequent lack of generalization for all cell types is considerably high.


\begin{figure}[!t]
 \centering
  \includegraphics[width=0.9\linewidth]{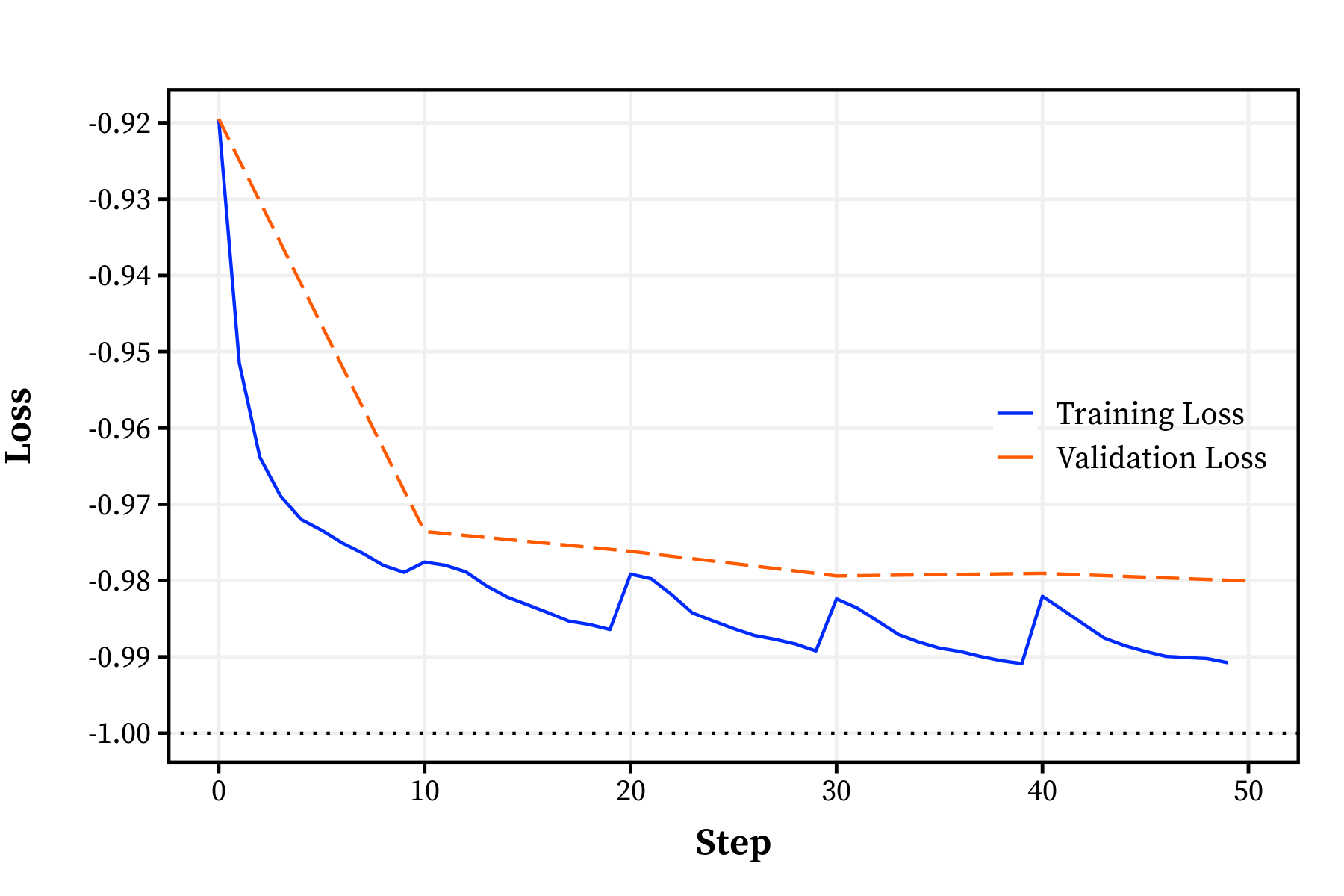}
  \caption{Segmentation loss: training and validation loss at each step (\textit{ETC: 32 minutes})}
  \label{fig:segmentation_loss}
 \end{figure}
 
First, the parameters of the loss function \textit{SSIM} \textit{(see \cref{ssim})} were defined. According to the work of \cite{SSIM} the parameters \textit{$k_{1}$} and \textit{$k_{2}$} have been maintained at their default values and only the size of the sliding window ($K$) has been modified, which makes the measurement sharper or softer. After reviewing the use of SSIM in different domains \citep{ssim1, ssim2} and analyzing its performance as a loss function in deep autoencoders \citep{segmentationStructural}, it was decided to set $K$-value to $5$. Even though this small value leads to a more complex learning process, the autoencoder carries enough structural capability to represent complex details such as busbar numbers, shapes, edges, and some other main features of the photovoltaic cell.

Instead of using pooling layers, strides were used since they offer better performance \citep{pooling}. \textit{LeakyRelu} \citep{leakyRelu} was selected as the activation function,  with an alpha value of $0.2$. The advantage of using this activation function in comparison with other functions is that \textit{LeakyRelu} does not require the use of batch normalization layers, thus keeping computational complexity lower. The dimensionality of latent space is very important, so a large value ($d=500$) was used to ensure the high-level features of the last convolutional layers keep persistent on the latent space.

\begin{figure}[!ht]
\centering

\subfloat[]{
\includegraphics[width=0.3\columnwidth]{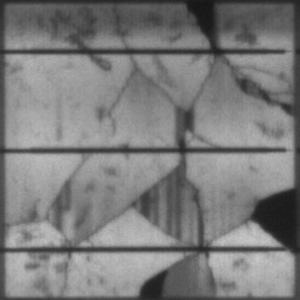}
\quad
\includegraphics[width=0.3\columnwidth]{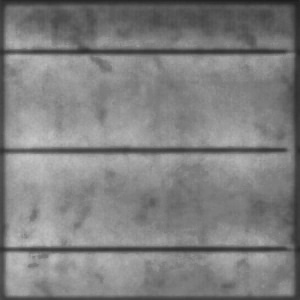}
\quad
\includegraphics[width=0.3\columnwidth]{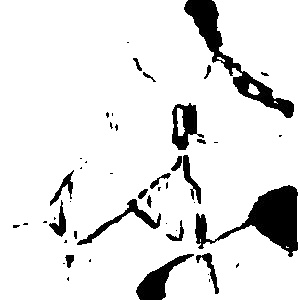}}\hfill

\subfloat[]{
\includegraphics[width=0.3\columnwidth]{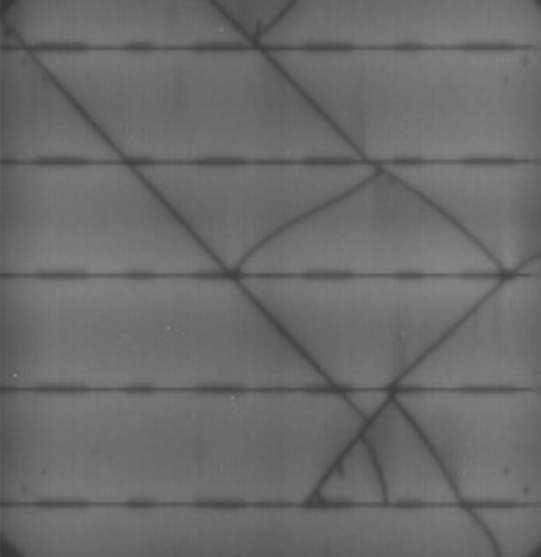}
\quad
\includegraphics[width=0.3\columnwidth]{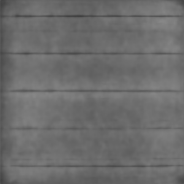}
\quad
\includegraphics[width=0.3\columnwidth]{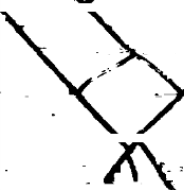}}\hfill


\subfloat[]{
\includegraphics[width=0.3\columnwidth]{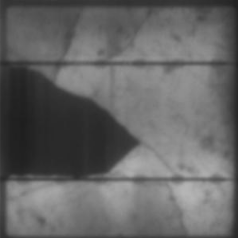}
\quad
\includegraphics[width=0.3\columnwidth]{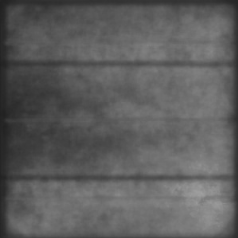}
\quad
\includegraphics[width=0.3\columnwidth]{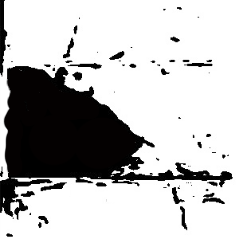}}\hfill

\subfloat[]{
\includegraphics[width=0.3\columnwidth]{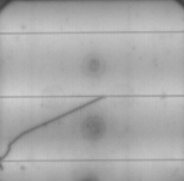}
\quad
\includegraphics[width=0.3\columnwidth]{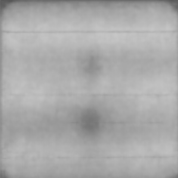}
\quad
\includegraphics[width=0.3\columnwidth]{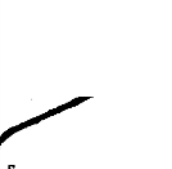}}\hfill

\caption{Illustration of the performance of the generative autoencoder and the segmentation algorithm \textit{(see \cref{alg:segmentation})}. Left: original image. Center: generated image. Right: difference between the original and the generated.}
\label{fig:example_segmentation}
\end{figure}

Once the parameter tuning has been concluded, a batch size of $8$ was established and trained the model for $13400$ steps, with a learning rate of $0.0015$. To measure the efficiency of the training, a validation process was performed at each epoch ($268$ steps). Since no validation set was available, it was decided to use k-folds cross-validation, with $k=10$. Hence, for each epoch, a random fold was selected among the $10$ available to perform the validation and the rest for training.
The evolution of the training loss \textit{(see \cref{fig:segmentation_loss})} shows that the loss has been decreasing uniformly converging in a value close to $0.992$ \textit{(see \cref{fig:example_segmentation})}.

The performance of the model is strongly correlated to the quality of the images. The finer the cracks, the better the image quality needed to capture all the anomalies. Furthermore, the post-processing algorithm uses adaptive thresholding, which determines the threshold for a pixel-based on a small region around it. This may cause the thinnest lines to be omitted in the mask generation. In addition, the model is trained using several types of cells, some of them generate stains and shades when capturing (e.g. \textit{polycrystalline} cells, see \cref{fig:example_segmentation}a). Having a single model to recreate different types of cells makes it less precise to slight anomalies.


Besides, the thinnest lines may be less prominent when there is a remarkably large anomaly in the same cell. In other words, if the cell has only fine cracks or dark areas, the detection may be done correctly (see \cref{fig:example_segmentation}b,\ref{fig:example_segmentation}d). However, when both thin lines and dark areas appear in the cell, the post-processing algorithm has difficulties binarizing the image precisely (see \cref{fig:example_segmentation}a,\ref{fig:example_segmentation}c). Moreover, the performance of the segmentation auto-encoder may also affect the quality of the generated cells, which can contain remaining anomalous features that may impact the post-processing algorithm.

\section{Conclusions and Discussions}
\label{section:conclusions_and_discussions}

In this work an efficient and automatized pipeline has been proposed, to detect and locate the finest defects of PV cells. The detection of such defects is a notable step to prolong the life of the panels and anticipate massive failures. Although this task is crucial, the current state-of-the-art detection methods barely extract low-level information from individual PV cell images, and their performance is conditioned by the available training data. The proposed end-to-end deep learning pipeline is able to detect, locate and segment cell level anomalies from entire photovoltaic panels via EL images \textit{(see \cref{fig:architecture_el})}. 

The modularity of the pipeline allows to design and evaluate the three deep learning models separately. The cell detector, with a precision of $99.36 \%$ (AP), shows that the model is highly reliable and robust. In the case of cell classifier, as mentioned above, our contribution outperforms the approach of \cite{deitsch_2018}. Although their model has slightly better accuracy ($0.863$), it has been trained and tested on a less complex dataset \citep{elpv1} with less PV cell variations, whilst our model has shown similar performance ($0.84$) in a more complex dataset on several types of PV cells. Finally, the anomaly segmentation model has obtained an accuracy of $0.992$ (SSIM) in the validation set. Despite the complexity of the weakly supervised technique and the lack of images, the obtained results show that this approach meets the expected performance. To the best of our knowledge, this is the first work that proposes a complete end-to-end solution for the detection and segmentation of cell-level anomalies in PV panels.

The anomaly segmentation module is a novel contribution for EL images that is capable of finding cracks, micro-cracks, dead spots, weak areas and weak cells. As weak regions end up dissipating some of the power generated by the more efficient cells, the detection of such defects is of vital importance for system integrators, panel manufacturers and cell fabricators. Therefore, a pipeline like the one here proposed can help to raise the quality standards of the PV panels production lines and can be a key tool for manufacturers to improve cell efficiency.

We plan to extend this work in several directions. First, the proposed pipeline has been tested on two datasets, with a broad spectrum of PV cell types and materials. In the event of new data with distinct domain features, the performance of the proposed method could be compromised, and the most vulnerable module would be the anomaly segmentation module, since it learns to generate images directly from the data distribution. Hence, we will further investigate a way to automatically annotate and integrate unseen data and further enhance the robustness of the model. Second, we intend to expand the pipeline with more functionalities. For instance, prior to the cell detection from PV panels, it could be useful to extract the panels from raw aerial images. In addition, a valuable extension to the anomaly segmentation module would be a classification module that collects more information about the type of anomaly. These two auxiliary modules would be subject to the availability of such data.

\section*{Acknowledgment}
\label{section:acknowledgment}

This publication resulted (in part) from the PROMISE (Advances in Photovoltaic Solar Energy Operation and Maintenance Research) project (KK2019/00088), which is financed by the ELKARTEK program of the Basque Government and is a collaborative project between Tecnalia, Tekniker, Vicomtech, UPV-ISG, UPV-TIM and Koniker. W. Cambarau would like to acknowledge I.Aizpurua, J.M. Calama and I.Arrizabalaga for contributing to the generation of the TecnaliaPR dataset.


\end{sloppypar}

\bibliography{references}

\end{document}